\documentclass[10pt,twocolumn,letterpaper]{article}

\usepackage[pagenumbers]{cvpr} 

\definecolor{cvprblue}{rgb}{0.21,0.49,0.74}
\usepackage[pagebackref,breaklinks,colorlinks,allcolors=cvprblue]{hyperref}
\usepackage[ruled,linesnumbered]{algorithm2e}
\usepackage{amssymb}
\usepackage{algorithmic}
\usepackage{graphicx}
\usepackage{textcomp}
\usepackage{xcolor}
\usepackage{booktabs}
\usepackage{framed}
\usepackage{tcolorbox}
\usepackage{multirow}
\usepackage[capitalize]{cleveref}
\usepackage{threeparttable}
\newcommand{\tool}[1]{\textsc{#1}\xspace}
\newcommand{\rt}{\tool{RedTest}}    

\def\paperID{2411} 
\def\confName{CVPR}
\def\confYear{2025}

\title{\rt: Towards Measuring Redundancy in Deep Neural Networks Effectively}

\author{
Yao Lu\textsuperscript{1}\thanks{yaolu.zjut@gmail.com} 
\quad Peixin Zhang\textsuperscript{2}
\quad Jingyi Wang\textsuperscript{3} 
\quad Lei Ma\textsuperscript{4,5} 
\quad Xiaoniu Yang\textsuperscript{1}
\quad Qi Xuan\textsuperscript{1}\thanks{Corresponding author: xuanqi@zjut.edu.cn, wangjyee@zju.edu.cn.}
\\
\textsuperscript{1}{Zhejiang University of Technology}
\quad \textsuperscript{2}{Singapore Management University} \\
\quad \textsuperscript{3}{Zhejiang University} 
\quad \textsuperscript{4}{The University of Tokyo} 
\quad \textsuperscript{5}{University of Alberta}
}

\begin{document}
\maketitle

\begin{abstract}
Deep learning has revolutionized computing in many real-world applications, arguably due to its remarkable performance and extreme convenience as an end-to-end solution. However, deep learning models can be costly to train and to use, especially for those large-scale models, making it necessary to \emph{optimize} the original overly complicated models into smaller ones in scenarios with limited resources such as mobile applications or simply for resource saving. 
The key question in such model optimization is, \emph{how can we effectively identify and measure the redundancy in a deep learning model structure}. While several common metrics exist in the popular model optimization techniques to measure the performance of models after optimization, they are not able to quantitatively inform the degree of remaining redundancy.
To address the problem, we present a novel testing approach, i.e., \rt (short for \emph{Red}undancy \emph{Test}ing), which proposes a novel testing metric called \emph{Model Structural Redundancy Score} (MSRS) to quantitatively measure the degree of redundancy in a deep learning model structure. We first show that MSRS is effective in both revealing and assessing the redundancy issues in many state-of-the-art models, which urgently calls for model optimization. Then, we utilize MSRS to assist deep learning model developers in two practical application scenarios: 1) in Neural Architecture Search, we design a novel redundancy-aware algorithm to guide the search for the optimal model structure (regarding a given task) and demonstrate its effectiveness by comparing it to existing standard NAS practice; 2) in the pruning of large-scale pre-trained models, we prune the redundant layers of pre-trained models with the guidance of layer similarity to derive less redundant ones of much smaller size. Extensive experimental results demonstrate that removing such redundancy has a negligible effect on the model utility.
\end{abstract}

\vspace{-6mm}    
\section{Introduction}
\label{sec:introduction}
Deep neural networks (DNNs) have achieved great success in many application scenarios including image understanding \cite{he2016deep}, object detection~\cite{redmon2016you} and autonomous vehicle~\cite{maqueda2018event}. However, such success comes at a price, as the complex training process and the power consumption, especially those large-scale state-of-the-art models, can be costly,
which requires significant computational and storage expense~\cite{DBLP:conf/naacl/DevlinCLT19,DBLP:conf/iclr/DosovitskiyB0WZ21}. For instance, on our experimental server, the storage space for ResNet152 with 60.2M parameters takes as much as 230.43 MB, and it would further take 226.06 MB memory and 11.56B Float Points Operations to infer the category of an image for the model. Such huge computation and storage consumption severely limits their deployment, especially on resource-constrained applications, e.g., mobile devices, or simply energy-critical scenarios. 

\begin{figure*}[t]
  \centering
   \includegraphics[width=0.9\linewidth]{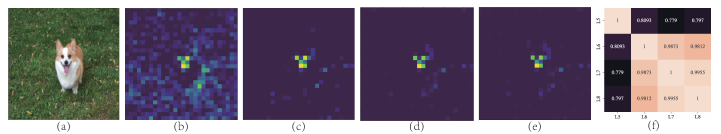}
   \caption{(a): Input image ($224 \times 224$). (b)-(e): Visualization of IRs ($28 \times 28$) in layers 5 to 8. (f): Similarity matrix. $L_i$ denotes i-th layer.}
   \label{fig:similar_redundancy}
\end{figure*}

To reduce the cost of deep learning applications, an intuitive way is to reduce the model size. In the ideal case, the model structure should be as simple as possible, to satisfy the utility requirement. 
Up to the present, we identify that existing research in machine learning community mostly follows the
the following two main lines to achieve such a goal: \textit{Model Pruning} (MP) and \textit{Neural Architecture Search} (NAS). MP tries to remove the redundant part in the model structure, at different levels including weights~\cite{han2015learning,liu2018frequency,zhang2018systematic,DBLP:conf/iclr/FrankleC19,chen2023rgp}, filters~\cite{wang2021accelerate,huang2018data,lin2020hrank,tang2021manifold,gao2021network,sui2021chip} and layers~\cite{chen2018shallowing,wang2021accelerate,huang2018data,lu2021graph,lu2024generic,tang2023sr}, based on their contributions to model inference. NAS adopts different search strategies, e.g., reinforcement learning~\cite{DBLP:conf/iclr/BakerGNR17,Zoph_2018_CVPR,Zhong_2018_CVPR,pmlr-v80-pham18a}, evolutionary algorithm~\cite{DBLP:conf/aaai/RealAHL19,pmlr-v70-real17a,DBLP:journals/corr/abs-2008-10937,DBLP:conf/iccv/XieY17} and gradient-based method~\cite{DBLP:conf/iclr/LiuSY19,DBLP:journals/corr/abs-1909-06035,DBLP:conf/iccv/ChenXW019}, to guide the automatic selection of an optimal model structure, which balances between model accuracy and model size, from a large number of candidate models. Arguably, either for MP or NAS, \emph{the key challenge in reducing the model size is to effectively identify the redundant parts in the model structure.} While existing MP and NAS techniques often leverage a set of metrics, e.g., the number of model parameters (Params)~\cite{chen2018shallowing,tan2019mnasnet}, the required Float Points Operations (FLOPs)~\cite{chen2018shallowing,lin2020hrank} and the time model takes to process one image (Latency)~\cite{tan2019mnasnet,wu2019fbnet}, etc., to characterize the effectiveness of such model optimization, 
these existing metrics unfortunately can only measure the model complexity from different perspectives~\cite{hu2021model}, but are not able to quantitatively measure the remaining redundancy. In other words, \emph{they cannot characterize how far we are in obtaining a model with minimum redundancy}.

Unfortunately, so far, there still lacks a common consensus on the definition of redundancy, not to mention how to measure it effectively.
To bridge this gap, in this paper, we first propose to leverage the similarity of intermediate representations (IRs) to reflect and characterize the structural redundancy of deep learning models. The motivation 
comes from an empirical and quantitative observation. As shown in~\cref{fig:similar_redundancy}, IRs learned by layer 6 to layer 8 of ResNet50~\cite{he2016deep} are similar (either visually or quantitatively), 
indicating that these layers might be redundant in extracting useful information. 
Furthermore, we propose a novel testing metric to quantify the structural redundancy degree of DNNs, named \textit{model structural redundancy score} (MSRS). 
MSRS can be used to calculate the similarity of any pair of IRs, even for those with different shapes (which can not be processed to calculate classical similarity metrics, e.g., cosine similarity and $L_p$-norm distance). Compared to existing metrics (e.g., Params, FLOPs and Latency used in model compression), MSRS is 
specially designed to assess and facilitate the identification of structural redundancy of models. 

Equipped with MSRS, we finally present a novel testing framework, i.e., \rt, and apply \rt in two practical application scenarios to assist deep learning developers in developing better models: searching for the optimal model from scratch with NAS and pruning redundant large-scale pre-trained models. 
Our extensive experimental results demonstrate that \rt is helpful in identifying a better model by removing redundancy. 

To summarize, we make the following contributions:
\begin{itemize}
\item[$\bullet$] We propose a novel testing metric called Model Structural Redundancy Score (MSRS) for deep learning models, which provides a quantitative assessment of the degree of redundancy in the model structure. With MSRS, we systematically explore and investigate the redundancy issue in state-of-the-art deep learning models and confirm the ubiquitous presence of model redundancy, sometimes even to a surprisingly high level, which urgently calls for model structure optimization.

\item[$\bullet$] We propose \rt, a \emph{Red}undancy \emph{Test}ing framework, which provides an end-to-end solution for optimizing the model structure. Specifically, we consider two practical application scenarios: designing a model from scratch and shrinking a redundant pre-trained model. 
Specifically, we propose a redundancy-aware NAS algorithm on the basis of MSRS to search for better models with less structural redundancy. For the pruning of large-scale models, we design a new layer pruning method with the guidance of layer similarity to remove the structural redundancy.

\item[$\bullet$] We release \rt together with all the experimental data publicly available\footnote{https://anonymous.4open.science/r/RedTest-0318/}, to benchmark and provide the basis for future study in this direction, which we believe to be important to contribute to a more energy-friendly deep learning ecosystem.
\end{itemize}
\section{Preliminary}
\label{sec:preliminary}



In this work, we consider two practical application scenarios for model optimization: MP and NAS. They share the same goal, i.e., balancing model accuracy and efficiency. However, they work in different scenarios: MP tries to shrink the model from a pre-trained one, leveraging both the architecture and parameters of the pre-trained model as prior knowledge, while NAS tries to search for a good model from scratch.

\noindent\textbf{MP} can be further categorized into weight pruning~\cite{han2015learning,liu2018frequency,zhang2018systematic,DBLP:conf/iclr/FrankleC19}, filter pruning~\cite{wang2021accelerate,huang2018data,lin2020hrank,tang2021manifold,gao2021network,sui2021chip}, and layer pruning~\cite{chen2018shallowing,wang2021accelerate,huang2018data,lu2021graph}. Weight pruning drops redundant weights to obtain sparse weight matrices, which has limited applications on general-purpose hardware~\cite{han2016eie}. Filter pruning seeks to remove unimportant filters according to specific metrics. 
However, FLOPs, Params, and Latency reduction by filter pruning are constrained by the original model’s depth. Hence, in this work, we focus on layer pruning, which eliminates entire redundant layers for better inference speedup.

\noindent\textbf{NAS} aims to search for the optimal architecture $\alpha$ for a given resource budget (e.g., latency) that achieves a desirable performance on the validation set among all candidate architectures $S$, which can be formulated as a bi-level optimization problem: 
\begin{equation}
\begin{array}{c}
    \mathop{min}\limits_{\alpha \in S} \mathcal{L}(\alpha, \theta_{\alpha}^*, \mathcal{D}_{val}) \\
    \text{ s.t. } \theta_{\alpha}^* = \mathop{min}\limits_{\theta_{\alpha}}\mathcal{L}(\alpha, \theta_{\alpha}, \mathcal{D}_{train}),
\end{array}
\label{eq:NAS}
\end{equation} 
where $\mathcal{L}$, $\theta_{\alpha}$, $\mathcal{D}_{train}$ and $\mathcal{D}_{val}$ denote the loss function (e.g., cross-entropy loss), the parameters of $\alpha$, the training set and the validation set, respectively. Once $\alpha$ is found, it will be re-trained on $\mathcal{D}_{train}$ (or $\mathcal{D}_{train} + \mathcal{D}_{val}$) and tested on the test set $\mathcal{D}_{test}$.
\section{Motivation and Research Objective}
\label{sec:Motivation and Research Objective}
In this section, we present the motivation of \rt along with its two practical application scenarios.
\subsection{Motivation} 
\label{sec:Motivation}
\begin{figure}[t]
  \centering
   \includegraphics[width=0.99\linewidth]{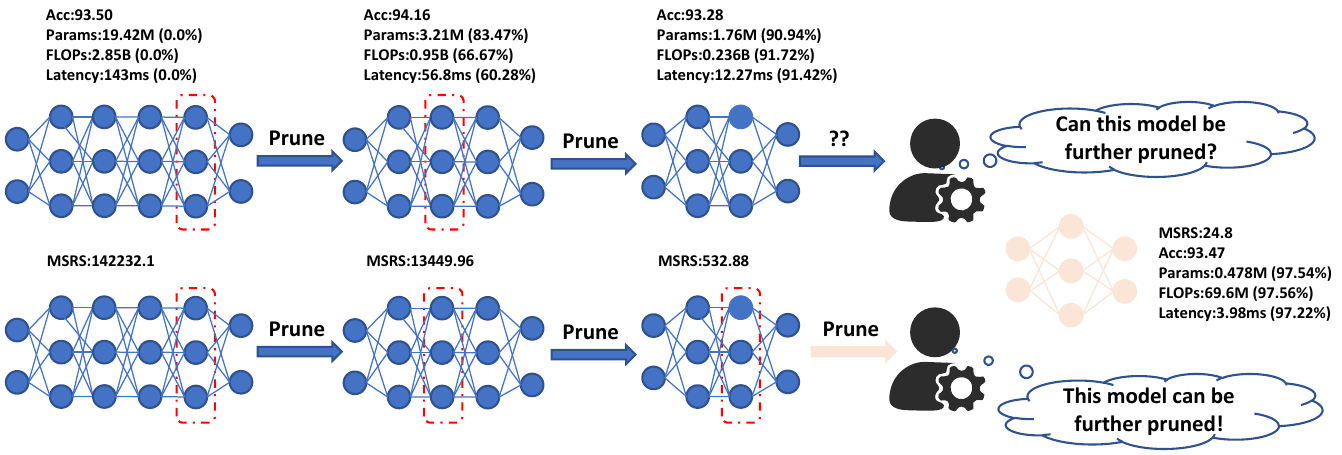}
   \caption{An intuitive example to explain the inadequacies of existing metrics.}
   \label{fig:motivation}
\end{figure}

The motivation for designing MSRS stems from the observation that 
existing metrics in measuring model optimization, e.g., Params, FLOPs and Latency can only
quantify the model complexity (after optimization), but are unable to quantitatively measure the structural redundancy in deep learning models. For ease of understanding, we take \cref{fig:motivation} as an example to intuitively illustrate the limitation of existing metrics in practice. 
We observe that the model complexity decreases and the model accuracy still maintains at an acceptable level as the model is gradually pruned. However, the developer can struggle to determine
whether the model can be further optimized without a measurement of the redundancy of the current model.
For example, in the extreme case of our experiments, even if the model complexity is reduced to 10\% of the original (i.e., 90.94\% parameters reduction, 91.72\% FLOPs reduction and 91.42\% Latency reduction), the model still yields 93.28\% top-1 accuracy, i.e., only with a slight accuracy drop. Our proposed redundancy measurement metric MSRS complements existing performance metrics and enables a developer to be aware of the model optimization progress.  

\subsection{Application Scenarios.} 
Nowadays, there are numerous well-designed models offered by PyTorch~\cite{DBLP:conf/nips/PaszkeGMLBCKLGA19}, TensorFlow~\cite{DBLP:conf/osdi/AbadiBCCDDDGIIK16} and third-party libraries~\cite{dong2021nats,DBLP:conf/emnlp/WolfDSCDMCRLFDS20}, or pre-trained models from open-source resources, e.g., OpenVINO\footnote{https://github.com/openvinotoolkit/open\_model\_zoo}. However, many of these models are overly complicated and can not be directly deployed on resource-constrained devices such as mobile or embedded devices due to the high computational cost. 

\rt is designed to work in any real-world setting where deep learning developers want a more lightweight model for suiting a specific task. 
It is worth noting that \rt is regarded as a static analysis tool where both the computational graph and the model parameters are fixed once trained.
Here, we provide the following two potential scenarios with numerical applications:
\noindent\begin{itemize}
\item[$\bullet$] \textbf{Application Scenario 1 (AS1):} Deep learning developers want to design suitable models from scratch to meet both the accuracy and computational requirement.
\item[$\bullet$] \textbf{Application Scenario 2 (AS2):} Deep learning developers want to prune large-scale pre-trained models without sacrificing too much accuracy to reduce as much resource consumption as possible.
\end{itemize}
\section{The Proposed \rt}
\label{sec:approach}
The core of our proposed \rt is a testing metric called MSRS to quantitatively measure model structural redundancy. Utilizing MSRS, we further provide two typical application scenarios: 1) searching for the optimal model from scratch, and 2) optimizing redundant pre-trained models.

\subsection{The MSRS Measurement}
\label{sec:Unbiased CKA}
The key idea of MSRS is to measure the similarities of IRs. The intuition is that the IRs store interesting information for the learning task and their mutual similarities can be useful to measure the redundancy in the model structure from an information extraction perspective. The higher the similarity is, the larger redundancy of the model structure is.

Note that since the output shapes of different hidden layers are various, it is impractical to utilize traditional metrics, such as cosine similarity and euclidean distance to directly measure the similarity between two IRs. 
We adopt Centered Kernel Alignment (CKA)~\cite{kornblith2019similarity} to address this challenge. Given two IRs, $F_i \in \mathbb{R}^{n \times \mathit{p}_i}$ and $F_j \in \mathbb{R}^{n \times \mathit{p}_j}$, one with $\mathit{p}_i$ neurons in $i$-th layer and another $\mathit{p}_j$ neurons in $j$-th layer, to the same $n$ samples, we measure the similarity between each pair of inputs through gram matrices $S_i = F_i F_i^T$ and $S_j = F_j F_j^T$. It is worth noting that the order of the gram matrix $S \in \mathbb{R}^{n \times n}$ is only related to $n$, i.e., it can avoid the problem of misalignment caused by different output shapes of hidden layers. Next, we adopt Hilbert-Schmidt Independence Criterion (HSIC)~\cite{gretton2005measuring} to calculate the statistical independence of $S_i$ and $S_j$. \begin{equation}
HSIC_0(S_i,S_j)=\frac{1}{(n-1)^2}tr(S_iHS_jH),
\label{eq:HSIC}
\end{equation}
where $H = I_n - \frac{1}{n}\textbf{1}\textbf{1}^T$ is the centering matrix. We refer readers to ~\cite{gretton2005measuring} for detailed derivation. However, $HSIC_0$ is not invariant to isotropic scaling, which is the basic requirement of similarity metric~\cite{kornblith2019similarity}. Thus, we further normalize $HSIC_0$ using \cref{eq:biased_CKA}:
\begin{equation}
CKA_0(S_i,S_j)=\frac{HSIC_0(S_i,S_j)}{\sqrt{HSIC_0(S_i,S_i)}\sqrt{HSIC_0(S_j,S_j)}}
\label{eq:biased_CKA}
\end{equation}
Another limitation of $HSIC_0(S_i,S_j)$ is that it has $O(\frac{1}{n})$ bias~\cite{gretton2005measuring}, where $n$ denotes the number of samples. In order to make the value independent of the batch size, we replace $HSIC_0(S_i,S_j)$ with an unbiased one~\cite{song2012feature}, which is depicted as follows:
\begin{equation}
    \begin{aligned}
    &HSIC_1(S_i,S_j) = \\
    &\frac{1}{n (n - 3)} \bigg[ tr (\tilde S_i \tilde S_j) + \frac{\textbf{1}^\top \tilde S_i \textbf{1} \textbf{1}^\top \tilde S_j \textbf{1}}{(n-1)(n-2)} - \frac{2}{n-2} \textbf{1}^\top \tilde S_i \tilde S_j \textbf{1} \bigg],
    \end{aligned}
\label{eq:unbias hsic}
\end{equation}
where $\tilde S_i$ and $\tilde S_j$ are derived from setting the diagonal entries of $S_i$ and $S_j$ to zero. Finally, unbiased CKA can be computed as follows:
\begin{equation}
CKA_1(S_i,S_j)=\frac{HSIC_1(S_i,S_j)}{\sqrt{HSIC_1(S_i,S_i)}\sqrt{HSIC_1(S_j,S_j)}},
\label{eq:unbiased_CKA}
\end{equation}
whose value ranges from 0 to 1 and a higher value is preferred.

\begin{figure}[t]
  \centering
   \includegraphics[width=1\linewidth]{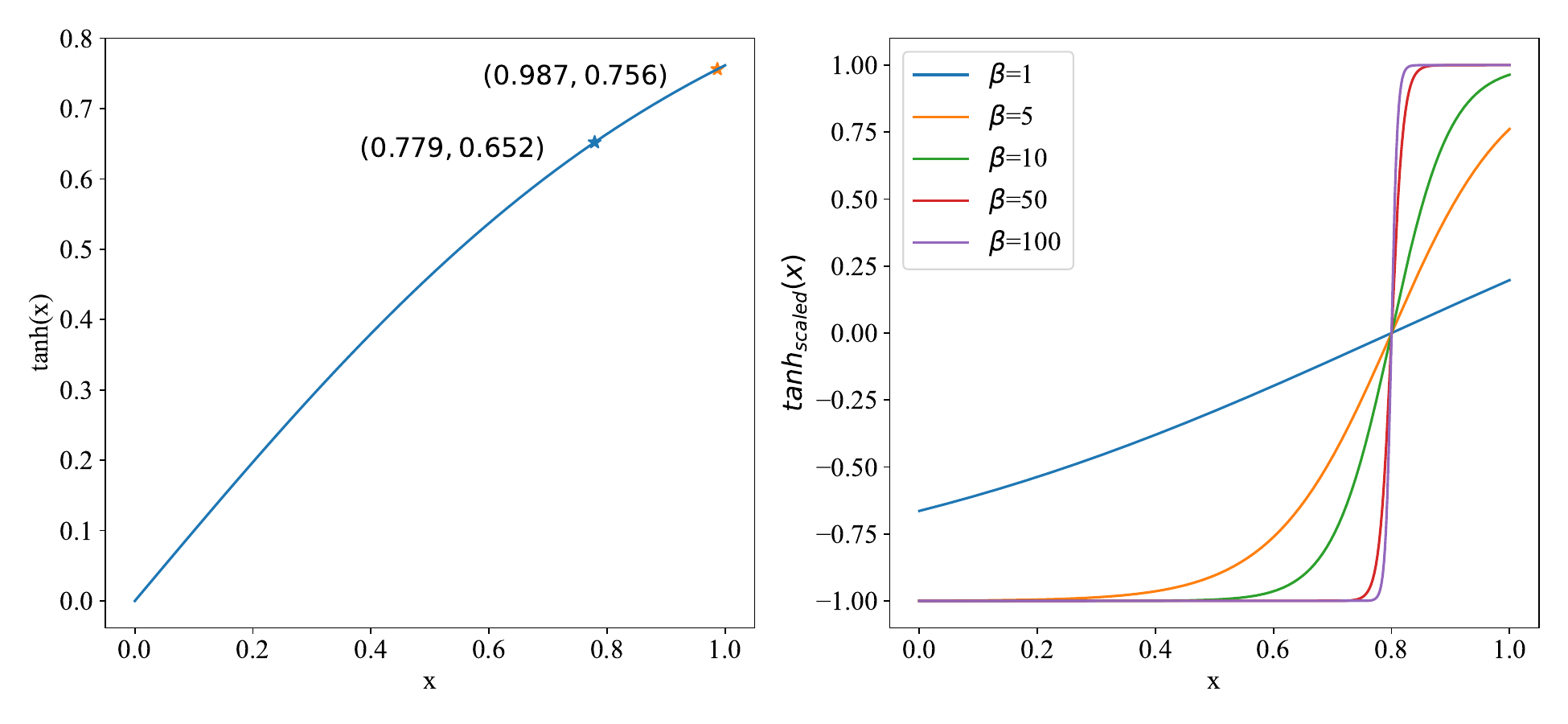}
   \caption{Representations of hyperbolic tangent function in $\left [ 0,1 \right ]$ and scaled hyperbolic tangent functions for $\epsilon = 0.8$ and various choices of $\beta$.}
   \label{fig:tanh}
  \vspace{-5mm}
\end{figure}

After obtaining the similarities between IRs, we further design a scoring mechanism to directly measure the redundancy degree.
As shown in~\cref{fig:similar_redundancy}, high similarity values reveal that the corresponding layers expose high redundancy. Hence, we prioritize high similarity values and assign higher weights to them. 
One widely adopted way in deep learning research is to use hyperbolic tangent function \cite{dubey2021comprehensive,kalman1992tanh}.
However, hyperbolic tangent function can not be directly applied here for two reasons: 1) The \cref{eq:unbiased_CKA} value is in $\left [ 0,1 \right ]$. If we use hyperbolic tangent function directly, those low similarity values have a non-negligible impact. For example, the assigned score (see the left plot of \cref{fig:tanh}) is $0.652$ for $x=0.779$, $0.756$ for $x=0.987$. Although the numerical difference between $0.652$ and $0.756$ is small, their similarity degrees are fundamentally different (see \cref{fig:similar_redundancy}). 2) It can be difficult to set a global similarity threshold to decide whether there are redundant layers because similarities between IRs may differ across various model structures in the real world. 

To address the problem, we adapt the original hyperbolic tangent function by introducing two hyperparameters: a scaling factor $\beta$ and a model-specific threshold $\epsilon$. 
\begin{equation}
tanh_{scaled}(x) = \frac{\exp(\beta (x - \epsilon)) - \exp(-\beta (x - \epsilon))}{\exp(\beta (x - \epsilon)) + \exp(-\beta (x - \epsilon))}
\label{eq:new_tanh}
\end{equation}
Specifically, the scaling factor $\beta$ controls the assigned score of the low similarity value. As $\beta$ increases, the effect of low similarity values on redundancy gradually decreases (see the right plot of \cref{fig:tanh}). When $\beta$ tends to infinity, \cref{eq:new_tanh} is equivalent to $sign(x - \epsilon)$. As for model-specific threshold $\epsilon$, we use various $\epsilon$ for different model structures. Empirical rules for picking $\beta$ and $\epsilon$ are summarized in RQ4.

\begin{algorithm}[t]
    \caption{The MSRS Measurement}
    \label{algorithm:MSRS}
    \LinesNumbered
    \KwIn {IRs set $\{F_1, F_2, \cdots, F_l\}$}
    \KwIn {model-specific threshold $\epsilon$}
    \KwIn {scaling factor $\beta$}
    Initialize MSRS=0; \\
    \For{i = 1 to l}{  \label{l1:loop_begin}
        $S_i = F_i F_i^T$ \\ \label{l1:gram1}
        \For{j = 1 to l}{ 
            $S_j = F_j F_j^T$ \\ \label{l1:gram2}
            $x=\frac{HSIC_1(S_i,S_j)}{\sqrt{HSIC_1(S_i,S_i)}\sqrt{HSIC_1(S_j,S_j)}}$ \\ \label{l1:cka}
            \If{$i > j$}{
                $score = \frac{\exp(\beta (x - \epsilon)) - \exp(-\beta (x - \epsilon))}{\exp(\beta (x - \epsilon)) + \exp(-\beta (x - \epsilon))}$ \\ \label{l1:score}
                $score = 0.5 \times score + 0.5$ \\ \label{l1:Normalization}
                $MSRS += score $ \label{l1:sum}
            }
      }
    } \label{l1:loop_end}
    \Return MSRS
\end{algorithm}

We provide a summary of calculating MSRS in Algorithm \ref{algorithm:MSRS}. Given a series of IRs $\{F_1, F_2, \cdots, F_l\}$, model-specific threshold $\epsilon$ and scaling factor $\beta$, we first initialize MSRS to 0. Then, we traverse IRs set twice to calculate gram matrices for the corresponding IRs (see line \ref{l1:gram1} and \ref{l1:gram2}). Next, we utilize each pair of gram matrices, i.e. $S_i$ and $S_j$, to calculate $CKA_{unbiased}$ (For simplicity, we use $x$ denotes $CKA_{unbiased}$) at line \ref{l1:cka}. Once the condition that $i > j$ is satisfied, we feed $x$ into the scaled hyperbolic tangent function to obtain a corresponding score (see line \ref{l1:score}) and normalize it (see line \ref{l1:Normalization}). Finally, we sum up all obtained scores at line \ref{l1:sum} to derive MSRS.

\begin{algorithm}[t]
    \caption{Redundancy-aware NAS}
    \label{algorithm:NASSelector}
    \KwIn {candidate architectures $\{m_1, m_2, \cdots, m_n\}$}
    \KwIn {expected MSRS $M$}
    \KwIn {expected Latency $T$}
    \KwIn {a tunable hyperparameter $\lambda$}
    \KwIn {weight factor $w$}
    Initialize $best\_model = \emptyset$, $best\_score = 0$ \\ \label{li:Initialize}
    \For{$i = 1$ to $n$}{
        Using Algorithm~\ref{algorithm:MSRS} to calculate the MSRS of $m_i$\\ \label{li:msrs}
        $score = ACC(m_i) \times \begin{bmatrix}\lambda \times \frac{LAT(m_i)}{T} + (1 - \lambda) \times \frac{MSRS(m_i)}{M} \end{bmatrix}^{w}$\\ \label{li:score}
        \If{$score > best\_score$}{ \label{li:start update}
            $best\_model = m_i$ \\
            $best\_score = score$ \\
            } \label{li:end update}
    }
    \Return $best\_model$ \label{li:best model}
\end{algorithm}

\subsection{AS1: Redundancy-aware NAS}
\label{sec:selection}

Designing redundancy-aware algorithms capable of automating the discovery of optimal architectures is critical for NAS to tradeoff between performance and the resource budget. Extensive prior research~\cite{tan2019mnasnet,wu2019fbnet,DBLP:conf/iclr/XieZLL19} has been explored 
in the direction of NAS to find a Pareto optimal solution by incorporating different metrics into the scoring function, e.g., Latency \cite{tan2019mnasnet} as follows.
\begin{equation}
\mathop{max}\limits_{m} \quad ACC(m) \times \begin{bmatrix}\frac{LAT(m)}{T} \end{bmatrix}^{w},
\label{eq:MnasNet}
\end{equation}
where $ACC(m)$, $LAT(m)$, $T$ and $w$ denote the accuracy of model $m$ on the specified dataset, the Latency of model $m$ on the specified device, expected Latency and application-specific constant, respectively. See \cite{tan2019mnasnet} for the empirical rule of thumb for picking $w$. 
In \cref{sec:Motivation}, we have clarified the insufficiency of existing metrics including Params, FLOPs, and Latency. To design redundancy-aware NAS, we propose to incorporate MSRS into the scoring function based on \cref{eq:MnasNet}. 
\begin{equation}
\mathop{max}\limits_{m} \quad ACC(m) \times \begin{bmatrix}\lambda \times \frac{LAT(m)}{T} + (1 - \lambda) \times \frac{MSRS(m)}{M} \end{bmatrix}^{w},
\label{eq:MSRS_select}
\end{equation}
where $MSRS(m)$, $M$ and $\lambda$ denote MSRS of model $m$, expected MSRS and a tunable hyperparameter. However, unlike expected Latency $T$ can be set according to real scenarios, it is difficult to propose a budget for expected MSRS $M$. To resolve this issue, we provide an empirical guide to picking expected MSRS in \cref{sec:budget}.



We integrate the designed score into the NAS algorithm (see in Algorithm \ref{algorithm:NASSelector}). 
To be specific, we first initialize the best model and score as $\emptyset$ and 0 in line \ref{li:Initialize}. Then, we traverse all candidate architectures, apply Algorithm~\ref{algorithm:MSRS} to calculate MSRS for each candidate architecture, and score it using \cref{eq:MSRS_select} (see line~\ref{li:msrs} and \ref{li:score}). If the current score is greater than the best score, we update the best model and score to the current model and score, respectively (see lines~\ref{li:start update} -~\ref{li:end update}). Finally, we obtain the model with the highest score in line~\ref{li:best model}.

\subsection{AS2: Redundancy-aware Layer Pruning}
\label{sec:pruning}
In \cref{sec:Unbiased CKA}, we obtain an MSRS for the target model, which reflects the structural redundancy in this model. For those highly redundant models, we design a brand new layer pruning method to optimize their model structures.

\begin{algorithm}[t]
    \caption{Redundancy-aware layer pruning}
    \label{algorithm:pruning}
    \LinesNumbered
    \KwIn {IRs set $\{F_1, F_2, \cdots, F_l\}$}
    \KwIn {similarity threshold $\mu$:}
    Initialize $remained\_layer = \emptyset$, $cur = i = 1$ \\
    \While{$i < l$}{ \label{li:loop_begin}
    	$i++$\\
    	$S_{cur} = F_{cur} F_{cur}^T$ \\ \label{li:cka_begin}
        $S_i = F_i F_i^T$ \\
    	$CKA_{unbiased}=\frac{HSIC_1(S_{cur},S_i)}{\sqrt{HSIC_1(S_{cur},S_{cur})}\sqrt{HSIC_1(S_i,S_i)}}$\\ \label{li:cka_end}
    	\If{$CKA_{unbiased} < \mu$}{
    		$remained\_layer = remained\_layer \cup cur$\\ \label{li:retain}
    	}
    	$cur = i$\\ \label{li:update_cur}
    } \label{li:loop_end}
    \Return remained\_layer \label{li:output}
\end{algorithm}

Different from existing layer pruning methods, our method builds upon the layer similarity between two adjacent layers. 
We prefer to retain the shallower layer of two adjacent layers to ensure the correctness of the prediction as much as possible.
For ease of understanding, we provide an example as follows. Let $F_{i-1}$, $F_i$ and $F_{i+1}$ be the outputs of three consecutive layers $L_{i-1}$, $L_i$ and $L_{i+1}$, respectively. We assume $F_i \sim F_{i+1}$, thus removing $L_{i+1}$ has less impact on subsequent inference of the model. On the contrary, if we discard $L_i$, then $F_{i-1}$ becomes the input of $L_{i+1}$ and the corresponding output now is $f_{i+1}(F_{i-1})$ ($f_{i+1}$ is the feature extraction function of $i+1$-th layer), which is totally different from $F_{i+1}$, since $F_{i-1}$ and $F_{i}$ are not similar.
Algorithm~\ref{algorithm:pruning} presents 
the details of our layer pruning method. We first initialize the remaining and current layers as $\emptyset$ and $1$, respectively. Next, we traverse IRs and calculate the layer similarity between the current layer and its next layer (see lines ~\ref{li:cka_begin}-\ref{li:cka_end}). If the unbiased CKA is smaller than the predefined similarity threshold $\mu$, we keep the current layer in the resultant model, otherwise, discard it (see line~\ref{li:retain}). Afterwards, we update the current layer to its next layer in line~\ref{li:update_cur} and continue the loop. Due to the layer drop, two consecutive layers in the resulting model may have dimensional mismatches. We thus modify the latter layer to match the front layer and finally obtain a model with as little redundancy as possible in line~\ref{li:output}. As for the compact model, we randomly initialize the parameters of those layers that changed dimensions and reuse the weights of pre-trained models for other layers. 
\section{Experimental Evaluation}
\label{sec:experiment}

We conduct experiments on four popular and publicly-available datasets including CIFAR10~\cite{krizhevsky2009learning}, CIFAR100~\cite{krizhevsky2009learning}, ImageNet~\cite{DBLP:journals/ijcv/RussakovskyDSKS15} and ImageNet16-120~\cite{chrabaszcz2017downsampled}. If not specified, we set the initial learning rate, batch size, weight decay, epoch and momentum to 0.01, 256, 0.005, 150 and 0.9, respectively. The learning rate is decayed by a factor of 0.1 at epochs 50 and 100. Besides, we utilize stochastic gradient descent algorithm to optimize the model. In total, we systematically measure the model structural redundancy over 35,000 DNN models to demonstrate the effectiveness of MSRS.


\begin{figure*}[htbp]
\centering
 \subfloat[]{
   \begin{minipage}[c]{0.23\textwidth}
     \centering
     \includegraphics[width=0.99\textwidth]{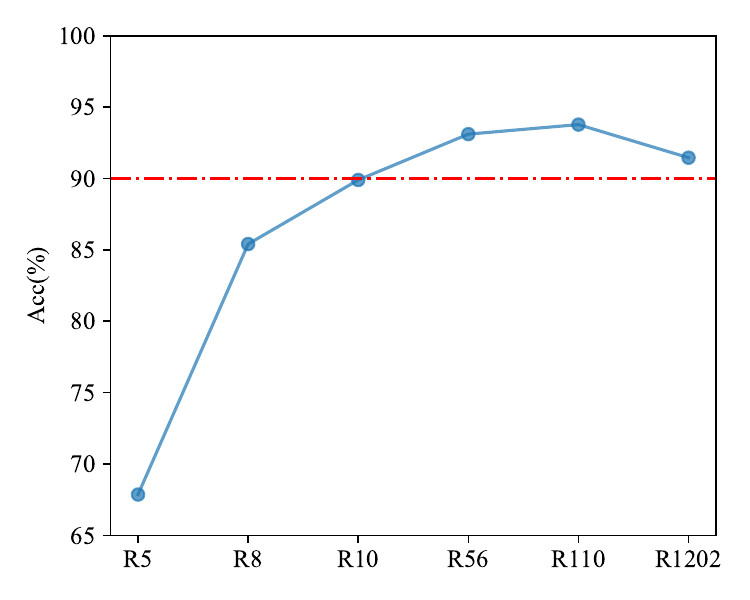}
   \end{minipage}
   \label{over_under_fitting1}
 }
 \subfloat[]{
   \begin{minipage}[c]{0.23\textwidth}
     \centering
     \includegraphics[width=0.99\textwidth]{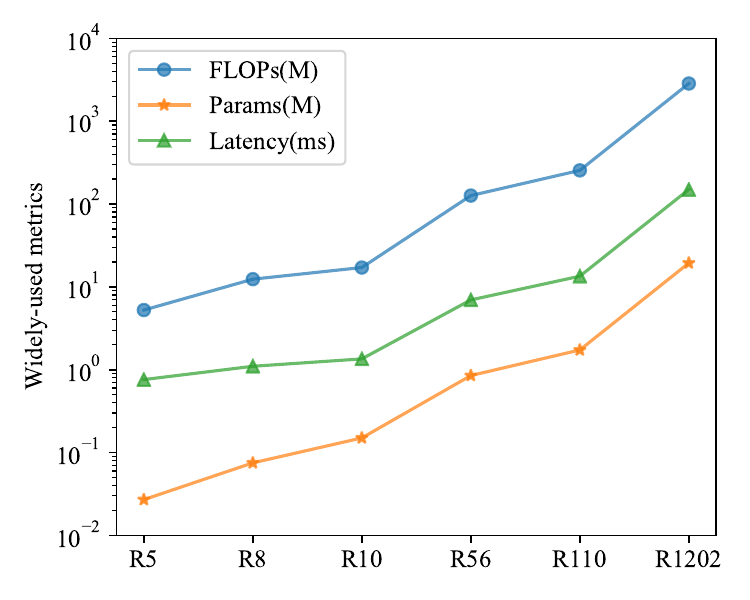}
   \end{minipage}
   \label{over_under_fitting2}
 }
 \subfloat[]{
   \begin{minipage}[c]{0.23\textwidth}
     \centering
     \includegraphics[width=0.99\textwidth]{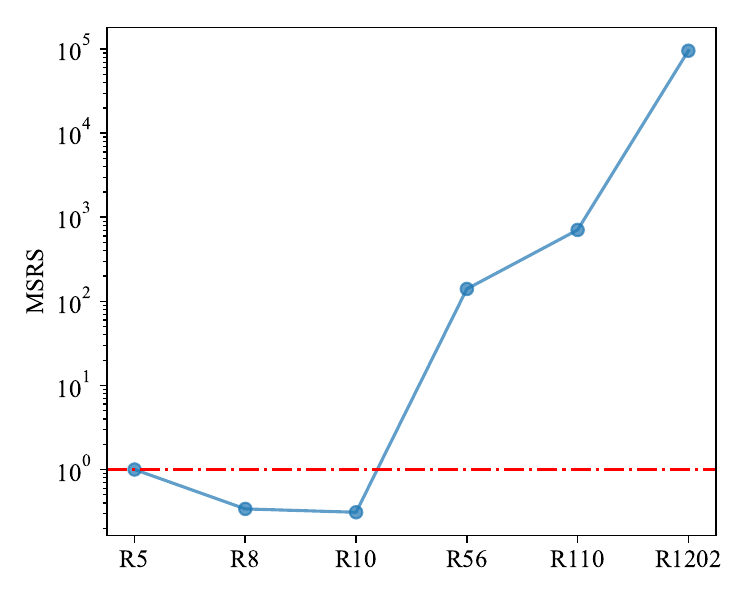}
   \end{minipage}
   \label{over_under_fitting3}
 }
\caption{Differences between existing metrics and MSRS for a family of ResNets on CIFAR10. R denotes ResNet.}
\label{over_under_fitting}
\vspace{-2mm}
\end{figure*}

\begin{figure*}[htbp]
  \centering
   \includegraphics[width=0.8\linewidth]{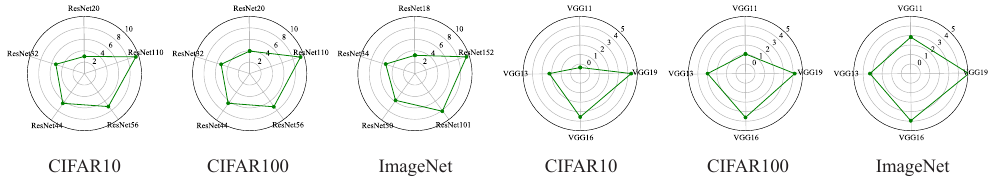}
   \caption{Structural redundancy in widely-used backbones on various datasets. To unify the coordinate axis scale, we take the binary logarithm of the calculated MSRS.}
   \label{fig:redundancy in backbones}
\end{figure*}

\label{sec:RQ}

\emph{\textbf{How effective is MSRS in measuring structural redundancy?}}
To answer the question, we compare MSRS with existing metrics, e.g., FLOPs, Params and Latency, in measuring a family of models whose redundancy increases gradually.
Specifically, we conduct experiments on the CIFAR10 and CIFAR100 dataset with a family of ResNets of different layers. Intuitively, as the 
number of layers increases, the ResNet model is transiting from underfitting to overfitting for the given learning task and the model redundancy is expected to grow at some point in the middle (where a model structure fits the task well).
As shown in \cref{over_under_fitting1}, ResNet5, ResNet8 and ResNet10 are still underfitting as indicated by their test accuracy, which have little redundancy. 
This matches their MSRS score well (all below 1 (see \cref{over_under_fitting3})).
Furthermore, as the model complexity grows from ResNet10 to ResNet56, the model redundancy does grow significantly from a certain point in the middle as we expected. For instance, when the model has over 1,000 layers, it tends to be overfitting and overparameterized ($93.78\%$ of ResNet110 vs. $91.46\%$ of ResNet1202). Meanwhile, the MSRS of ResNet1202 becomes extremely high, i.e., 95,817. 
These observations confirm that
our designed MSRS is strongly correlated with the model structural redundancy.
In contrast, FLOPs, Params and Latency are simply proportional to the model depth (see \cref{over_under_fitting2}), which is ineffective in revealing the structural redundancy issues. We can draw the same conclusion on CIFAR100 (See \cref{fig:over_under_fitting_cifar100} in Appendix for space reasons).


We further explore the structural redundancy in widely used ResNets~\cite{he2016deep} and VGGs~\cite{DBLP:journals/corr/SimonyanZ14a} models for image classification.
For the ImageNet dataset, we utilize pre-trained models from torchvision\footnote{https://pytorch.org/vision/stable/index.html} to calculate MSRS. For CIFAR10 and CIFAR100, we train models from scratch. The performance of the above models is summarized in \cref{Models statistics} of Appendix.
To unify the coordinate axis scale, we take the binary logarithm of the calculated MSRS. The results are shown in \cref{fig:redundancy in backbones}. We could observe that as the model (for both ResNets and VGGs) goes deeper, the degree of structural redundancy increases accordingly
for CIFAR10, CIFAR100 and ImageNet. A closer look reveals that although many models achieve comparable performance, their MSRSs vary significantly. MSRS provides a new dimension to evaluate and further optimize the model quality.
For instance, the model accuracy of ResNet56 and ResNet110 on CIFAR10 are almost the same ($93.11\%$ vs. $93.78\%$), while their MSRSs are quite different ($140.77$ vs. $708.12$). The underlying reason is that there are certain layers in ResNet110 extracting similar features leading to structural redundancy in the model.

\begin{tcolorbox}
\noindent \textbf{Summary:} MSRS is effective in measuring and revealing model structural redundancy, which is present (sometimes to a surprisingly high level) in many popular models.
\end{tcolorbox}

\emph{\textbf{How useful is MSRS in assisting NAS for the search of models with appropriate structure? 
}} To answer this question, we conduct experiments on NATS-Bench~\cite{dong2021nats}, a relatively fair and unified benchmark for the comparison of different NAS algorithms. With the provision of extra statistics, such as accuracy during training and Latency, running NAS algorithms on NATS-Bench can avoid the expensive computational overhead of training the architecture via directly querying from the database. Herein, we compare our redundancy-aware NAS algorithm (S3) with another two algorithms to evaluate its validity: S1~\cite{DBLP:conf/iclr/ZophL17,DBLP:conf/aaai/RealAHL19,DBLP:conf/iclr/LiuSY19} focuses on achieving higher accuracy, regardless of the associated resource budget, e.g., FLOPs, Params and Latency. S2~\cite{tan2019mnasnet} investigates a resource-aware NAS algorithm to tradeoff accuracy against resource consumption (Latency). 

\begin{figure}[t]
  \centering
   \includegraphics[width=1\linewidth]{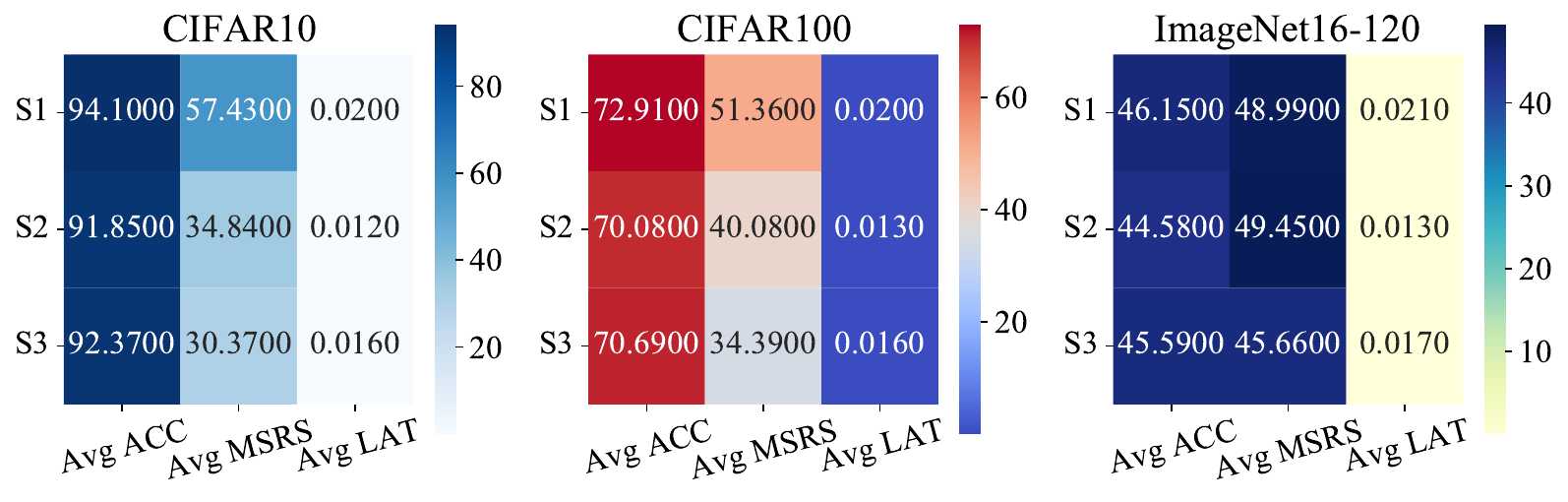}
   \caption{Statistics of top-ranking models in NATS-Bench selected by different strategies. Avg ACC (\%), Avg MSRS and Avg LAT (s) represent average accuracy, average MSRS and average Latency.}
   \label{fig:avg nas}
  \vspace{-5mm}
\end{figure}

Specifically, we respectively sample as many as $11,715$, $11,748$ and $11,741$ valid architecture candidates on CIFAR10, CIFAR100 and ImageNet16-120 on NATS-Bench with a topology search space and utilize Algorithm~\ref{algorithm:NASSelector} to select $1 \text{\textperthousand}$ top-ranking models (See \cref{tab:addtional hyper} for detailed parameter settings).  To be fair, we use the same $w$ and $T$ for S2 and our redundancy-aware NAS algorithm. Since final accuracy can be impractical to evaluate the relative ranking of such a large number of architectures, we use the accuracy of the 12-th epoch to approximate the final accuracy of each candidate~\cite{dong2021nats}, which allows us to perform architecture search quickly. 

Then, we calculate the average value of accuracy, MSRS and Latency of top-ranking models selected by various NAS algorithms (see \cref{fig:avg nas}). According to the experimental results, we observe that models selected by S1 have the highest average value on accuracy (94.10\% on CIFAR10, 72.91\% on CIFAR100 and 46.15\% on ImageNet16-120), Latency (0.02 on CIFAR10, CIFAR100 and 0.021 on ImageNet16-120) and MSRS (57.43 on CIFAR10, 51.36 on CIFAR100 and 48.99 on ImageNet16-120), which is attributed to the fact that S1 only focuses on accuracy but ignores the accompanying resource overhead. By contrast, S2 seeks to find the Pareto-optimal solutions to achieve accuracy-Latency trade-offs. Hence, the models selected by S2 can achieve lower Latency but exhibit an accuracy drop compared to S1. Specifically, the average accuracies and Latencies obtained by S2 are 91.85\% and 0.012 on CIFAR10, 70.08\% and 0.013 on CIFAR100 as well as 44.58\% and 0.013 on ImageNet16-120. Besides, we observe that S2 can eliminate model structural redundancy to a certain extent. However, since Latency cannot accurately reflect the structural redundancy existing in the models, models selected by S2 exhibit higher MSRS than S3 (our redundancy-aware NAS algorithm). Specifically, in comparison with S2, models selected by S3 have less structural redundancy (30.37 vs. 34.84 on CIFAR10, 34.39 vs. 40.08 on CIFAR100 and 45.66 vs. 49.45 on ImageNet16-120) while yielding higher accuracy (92.37\% vs. 91.85\% on CIFAR10, 70.69\% vs. 70.08\% on CIFAR100 and 45.59\% vs. 44.58\% on ImageNet16-120). Note that, the effectiveness of MSRS-assisted NAS in finding better models (with overall $~$30000 models evaluated) further confirms the usefulness of MSRS in measuring the redundancy. 
In \cref{sec:Additional experiments on redundancy-aware NAS}, we further incorporate Params and FLOPs into the scoring function to evaluate the effectiveness of the redundancy-aware NAS algorithm.
\begin{tcolorbox}
\noindent \textbf{Summary:} Compared to existing standard practice, our redundancy-aware NAS algorithm with the assistance of MSRS is able to find better models with less redundancy while sacrificing as little accuracy as possible.  
\end{tcolorbox}

\emph{\textbf{How effective is redundancy-aware layer pruning in reducing the model's structural redundancy?}}
We answer this question from two aspects. First, we conduct experiments on pruning with the guidance of layer similarity to evaluate the effectiveness of our redundancy-aware layer pruning in alleviating the model structural redundancy. Specifically, we utilize Algorithm~\ref{algorithm:pruning} to obtain models with little redundancy and then fine-tune them. All the hyperparameter settings are the same as normal training. For each model, we run three times and report the mean and standard deviation to avoid the effect of randomness. As an example, we visualize the results of CKA diagnosis on CIFAR10 with VGG19 and ResNet56 in \cref{fig:CKA pruning} of Appendix. Layers with high CKA values are labeled with a transparent color and are considered redundant. To evaluate the effectiveness of our method, we compare the performance with existing state-of-the-art layer pruning methods~\cite{lu2021graph,chen2018shallowing,wang2019dbp} in terms of accuracy and FLOPs/Params/Latency reduction. We re-implemented these baselines following the guidance and configurations of the original papers. We use these methods to select non-redundant layers on the same model and fine-tune the resulting model with the same setting as ours. For a fair comparison, we do not use the performance recovery strategies, such as knowledge distillation in \cite{chen2018shallowing, wang2019dbp} and iterative pruning in \cite{wang2019dbp}.

\cref{tab:CKA pruning} summarizes the experimental results of VGG19 and ResNet56 on CIFAR10. Generally, our redundancy-aware layer pruning surpasses the baselines. More specifically, as for ResNet56, 52.60\% FLOPs, 45.00\% parameters and 50.78\% Latency are removed by us, while it still yields 93.40\% top-1 accuracy. Compared with Chen et al.\cite{chen2018shallowing} and DBP\cite{wang2019dbp}, our method achieves better accuracy, FLOPs and Latency reduction, while maintaining the same parameters reduction. Note that the difference between Chen et al. and DBP lies in the techniques to recover the performance. Besides, although our method is slightly inferior to Lu et al.~\cite{lu2021graph} in FLOPs and Latency reduction, our method maintains better parameters reduction and better accuracy. As for VGG19, our method surpasses Lu et al. in all aspects, including top-1 accuracy, as well as FLOPs, parameters and Latency reduction (93.01\% vs. 91.7\% for accuracy, 54.41\% vs. 35.51\% for FLOPs, 63.34\% vs. 46.82\% for parameters and 40.16\% vs. 31.10\% for Latency). Finally, in comparison with Chen et al. and DBP, our redundancy-aware layer pruning has better accuracy, FLOPs and Latency reduction, while Chen et al. and DBP show more advantages in reduction of parameters. Furthermore, our method also surpasses the baselines on CIFAR100 (See \cref{tab:CKA pruning on cifar100} in Appendix for space reasons). Hence, our method works well in alleviating the model structural redundancy.
The above experiments in turn confirm that these high-similarity layers are redundant and can be pruned with minimal impact on model accuracy with the guidance of MSRS.
\begin{table}[htbp]
\centering
\scalebox{0.5}{
\begin{tabular}{cccccc}
\toprule
\textbf{Model} & \textbf{Method} & \textbf{Top-1\%} & \textbf{Params (PR)} & \textbf{FLOPs (PR)} & \textbf{Latency (PR)} \\
\midrule
\multirow{4}{*}{\textbf{ResNet56}} 
& Original & 93.11 & 0\% & 0\% & 0\% \\
& Chen et al.~\cite{chen2018shallowing}/DBP~\cite{wang2019dbp} & 92.79 $\pm$ 0.13 & 45.00\% & 41.30\% & 38.76\% \\
& Lu et al.~\cite{lu2021graph} & 92.7 $\pm$ 0.04 & 42.91\% & 60.23\% & 55.87\% \\
& Ours & \textbf{93.40} $\pm$ 0.05 & 45.00\% & 52.60\% & 50.78\% \\
\midrule
\multirow{4}{*}{\textbf{VGG19}} 
& Original & 93.36 & 0\% & 0\% & 0\% \\
& Chen et al.~\cite{chen2018shallowing}/DBP~\cite{wang2019dbp} & 92.86 $\pm$ 0.17 & 69.62\% & 44.92\% & 35.83\% \\
& Lu et al.~\cite{lu2021graph} & 91.71 $\pm$ 0.13 & 46.82\% & 35.51\% & 31.10\% \\
& Ours & \textbf{93.01} $\pm$ 0.17 & 63.34\% & 54.41\% & 40.16\% \\
\bottomrule
\end{tabular}}
\caption{Comparison of different layer methods.}
\label{tab:CKA pruning}
\vspace{-6mm}
\end{table}

After demonstrating the effectiveness of our redundancy-aware layer pruning, we further investigate structural redundancy in pruned models. Hence, we calculate the MSRS for pruned models obtained by various layer pruning methods and measure the corresponding MSRS reduction compared with original models. The results are shown in \cref{fig:msrs_before_after_pruning}, from which we find that the model structural redundancy is mitigated to varying degrees. Specifically, the original MSRS of ResNet56 and VGG19 on CIFAR10 are $140.77$ and $20.39$. Model structural redundancy completely disappears after applying layer pruning to ResNet56 (MSRS reduction is $100\%$ for all methods). By contrast, although the effect of mitigating model structural redundancy on VGG19 is not as good as on ResNet56, it also fully demonstrates the capability of existing layer pruning methods in reducing the structural redundancy of deep learning models (The MSRS reductions of Chen et al. and DBP, Lu et al. and ours are $91.5\%$, $73.3\%$ and $98.4\%$, respectively). Among these methods, our method achieves the best results, which further demonstrates the effectiveness of our redundancy-aware layer pruning.
\begin{tcolorbox}
\noindent \textbf{Summary:} Layer pruning is helpful for alleviating model structural redundancy to varying degrees, especially effective for models with high levels of redundancy. In addition, experiments demonstrate the effectiveness of our redundancy-aware layer pruning in reducing the model complexity.
\end{tcolorbox}

\section{Conclusion}
\label{sec:conclusion}
In this work, we develop \rt, a novel \emph{Red}undancy \emph{Test}ing framework, to measure the deep learning model's structural redundancy and help model developers to optimize the model structure. 
We first show that MSRS is effective in both revealing and assessing the redundancy issues in many state-of-the-art backbones, which urgently call for model optimization. Then, we further propose model optimization techniques to help assist model developers in two practical application scenarios: 1) in NAS, we propose a redundancy-aware NAS algorithm to guide the search for the optimal structure and show its effectiveness by comparing it to existing standard NAS practice; 2) in MP, we design a redundancy-aware layer pruning with the guidance of MSRS to prune the redundant layers of pre-trained models. Extensive experiments demonstrate that removing such redundancy has a negligible effect on the model utility. 
We hope that \rt can facilitate a much larger community of researchers to concentrate on optimizing model structures in a more computationally effective environment and ultimately create a greener deep learning ecosystem.

{
    \small
    \bibliographystyle{ieeenat_fullname}
    \bibliography{main}
}
\clearpage

\setcounter{figure}{0}
\setcounter{table}{0}
\renewcommand\thefigure{\Alph{section}\arabic{figure}} 
\renewcommand\thetable{\Alph{section}\arabic{table}}   
\appendix

\section{Additional Experiments on Redundancy-aware NAS}
\label{sec:Additional experiments on redundancy-aware NAS}

In this section, we incorporate Params and FLOPs instead of Latency into the scoring function and depict the scoring function as follows:
$$\mathop{max}\limits_{m} \quad ACC(m) \times \begin{bmatrix}\lambda \times \frac{Para(m)}{P} + (1 - \lambda) \times \frac{MSRS(m)}{M} \end{bmatrix}^{w},$$$$\mathop{max}\limits_{m} \quad ACC(m) \times \begin{bmatrix}\lambda \times \frac{FLOP(m)}{F} + (1 - \lambda) \times \frac{MSRS(m)}{M} \end{bmatrix}^{w},$$
where $Para(m)$, $P$, $FLOP(m)$ and $F$ denote Params of model $m$, expected Params, FLOPs of model $m$ and expected FLOPs, respectively. Meanwhile, the scoring functions of S2 are formalized as follows:
$$\mathop{max}\limits_{m} \quad ACC(m) \times \begin{bmatrix}\frac{Para(m)}{P} \end{bmatrix}^{w},$$$$\mathop{max}\limits_{m} \quad ACC(m) \times \begin{bmatrix}\frac{FLOP(m)}{F} \end{bmatrix}^{w},$$

\begin{figure*}[htbp]
\centering
 \subfloat[]{
   \begin{minipage}[c]{0.5\textwidth}
     \centering
     \includegraphics[width=0.99\textwidth]{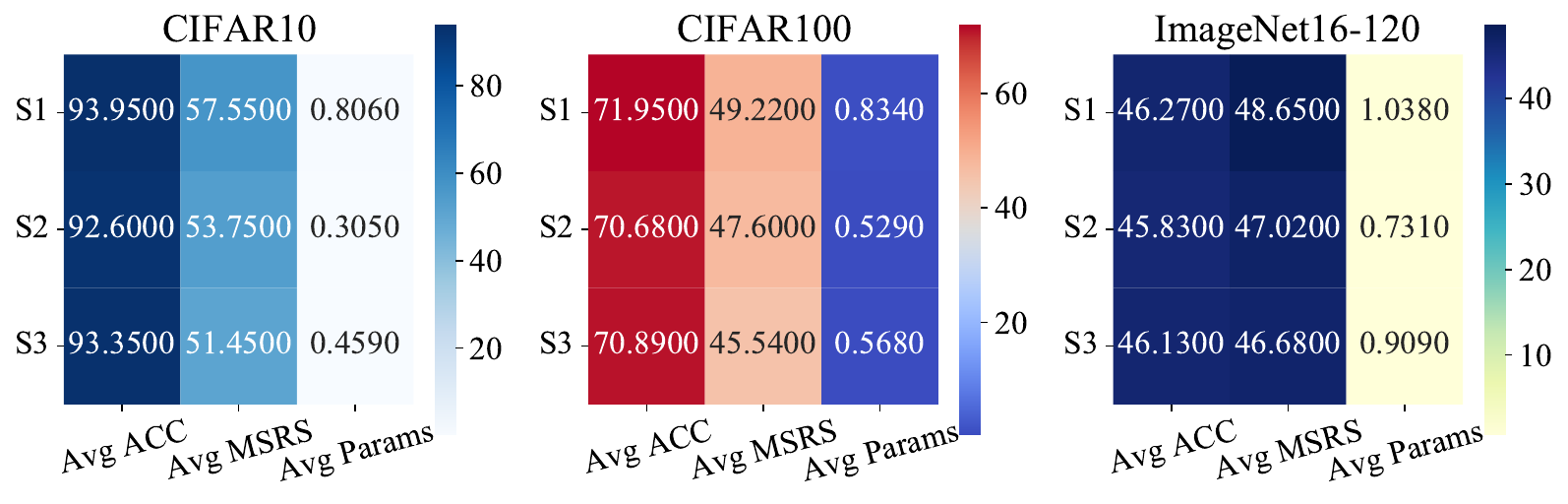}
   \end{minipage}
   \label{fig:avg_nas_params}
 }%
 \subfloat[]{
   \begin{minipage}[c]{0.5\textwidth}
     \centering
     \includegraphics[width=0.99\textwidth]{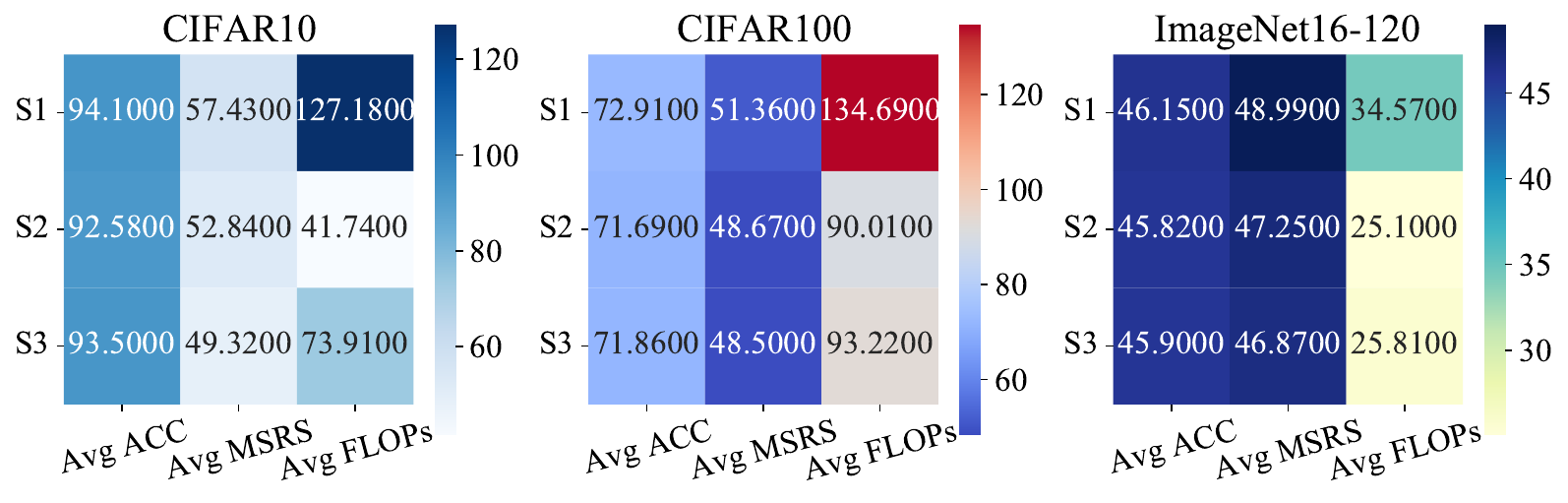}
   \end{minipage}
   \label{fig:avg_nas_flops}
 }
\caption{Statistics of top-ranking models. Avg ACC (\%), Avg MSRS, Avg Params (M), and Avg FLOPs (M) represent average accuracy, average MSRS, average Params, and average FLOPs, respectively.}
\label{fig:nas_selector_final}
\vspace{-5mm}
\end{figure*}

Specifically, we respectively sample $11715$, $11748$ and $11741$ valid architecture candidates on CIFAR10, CIFAR100 and ImageNet16-120 in the topology search space of NATS-Bench and utilize Algorithm~\ref{algorithm:NASSelector} to select top-ranking models. As for redundancy-aware NAS using FLOPs, we select $1 \text{\textperthousand}$ top-ranking models. As for redundancy-aware NAS using Params, we select $5 \text{\textperthousand}$ top-ranking models on CIFAR10 and CIFAR100, $0.5 \text{\textperthousand}$ top-ranking models on ImageNet16-120. \cref{tab:addtional hyper for Params} and \cref{tab:addtional hyper for FLOPs} provide the detailed experimental parameter settings. As shown in \cref{fig:nas_selector_final}, we can draw the same conclusion that our redundancy-aware NAS is proficient at selecting models with less redundancy and better performance.

\section{How to pick expected MSRS}
\label{sec:budget}
\begin{figure}[htbp]
  \centering
   \includegraphics[width=0.5\linewidth]{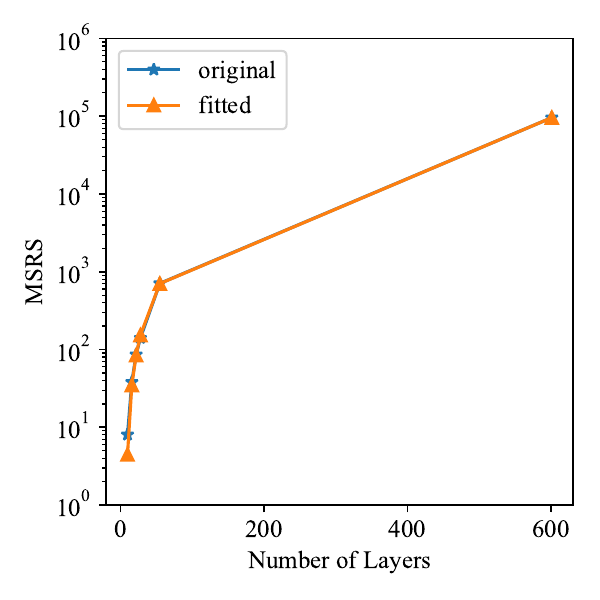}
   \caption{The relationship between the number of layers and MSRS.}
   \label{fig:fitting}
  \vspace{-3mm}
\end{figure}

\begin{figure*}[htbp]
  \centering
   \includegraphics[width=1\linewidth]{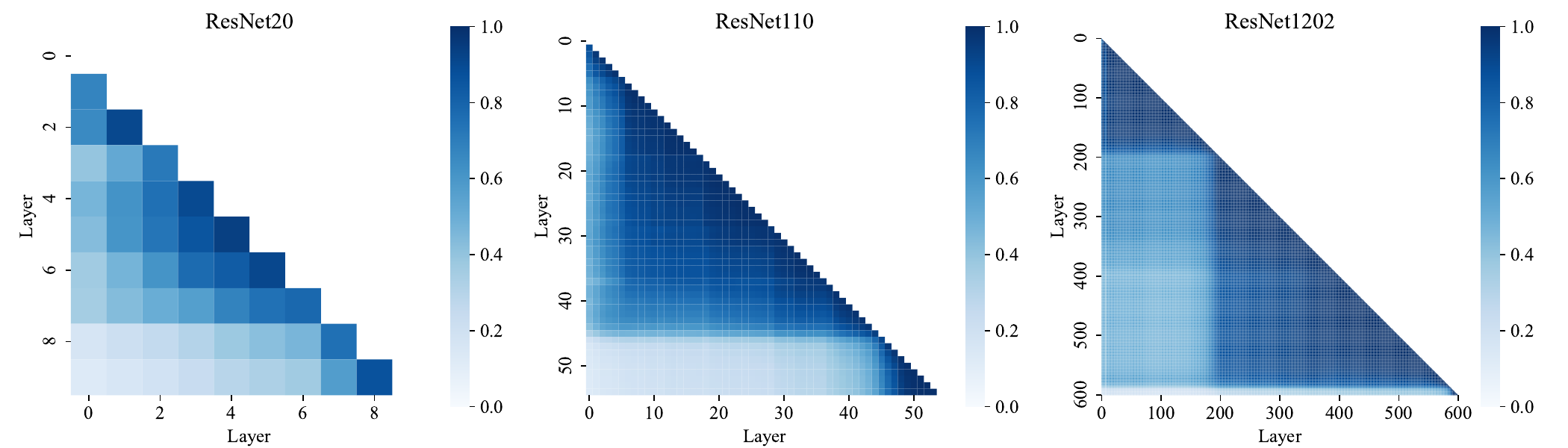}
   \caption{The lower triangular similarity matrices of ResNets on CIFAR10. The darker the color, the more similar.}
   \label{fig:budget}
\end{figure*}

In this section, we elaborate on the empirical guide to picking expected MSRS as a budget. Before that, we first explore the similarity matrices of ResNets on CIFAR10 (see \cref{fig:budget}). We observe that values with high similarity tend to take on a triangular shape as layers deepens. We believe that is because many successive layers are essentially repeating the same operation multiple times, refining features just a little more each time. Hence, these successive layers are highly similar, resulting in the triangular shape for high similarity values in similarity matrices. On the basis of this finding, we assume that the redundancy is quadratic with number of layers (depth) of the given model. In order to further verify our conjecture, we fit a quadratic function of depth with respect to MSRS according to the commonly used ResNets, which is shown in \cref{fig:fitting}. We find that the fitted curve is almost coincides with the original curve, which confirms our conjecture. Afterwards, we revisit the similarity matrix of shallow models. We find that values with high similarity are rare and show a scatter-like distribution (see the left plot of \cref{fig:budget}). Hence, we boldly guess that MSRS should be proportional to the layer depth. Next we empirically pick up multiple low-redundancy models with different depths and fit linear functions on CIFAR10 and CIFAR100, which are formalized as belows:
$$M_{C10} = \begin{cases} 0&\text{$l \leq$ 14} \\ 7.6 \times l - 106.4& \text{$l ~\textgreater$ 14}\end{cases},$$ $$M_{C100} = \begin{cases} 0&\text{$l \leq$ 3} \\ 2.4 \times l - 8.27& \text{$l ~\textgreater$ 3}\end{cases},$$ where $l$ denotes the model depth. C10 and C100 are CIFAR10 and CIFAR100.

\section{Experimental setup}
\subsection{Dataset Description}
CIFAR10 and CIFAR100 consist of 50K training images and 10K testing images in 10 and 100 fine-grained classes, respectively. ImageNet spans 1,000 classes and contains 1.28M training images and 50K validation images. ImageNet16-120 consists of 151.7K training images, 3K validation images, and 3K testing images, whose images are down-sampled from the original ImageNet to $16 \times 16$ pixels and selected from ImageNet with label $\in [1, 120]$. \cref{tab:datasets} in Appendix provides a brief description of the datasets statistics.

\subsection{Hyperparameter Settings} If not specified, we utilize minibatch of size $n = 256$ and $\beta=100$ to calculate MSRS. As for model-specific threshold $\epsilon$, we set $\epsilon=0.7$ for models of plain structure (e.g., VGGs) and $\epsilon=0.8$ for models of block structure (e.g., ResNets). More discussions on the selection of these configurations are presented in \cref{Hyperparameter selection of MSRS}. Besides, for calculating MSRS, we repeat each configuration 10 times and report the averaged results to counteract the effects potentially brought by randomness.

\begin{figure}[htbp]
\centering
 \subfloat[]{
   \begin{minipage}[c]{0.21\textwidth}
     \centering
     \includegraphics[width=1\textwidth]{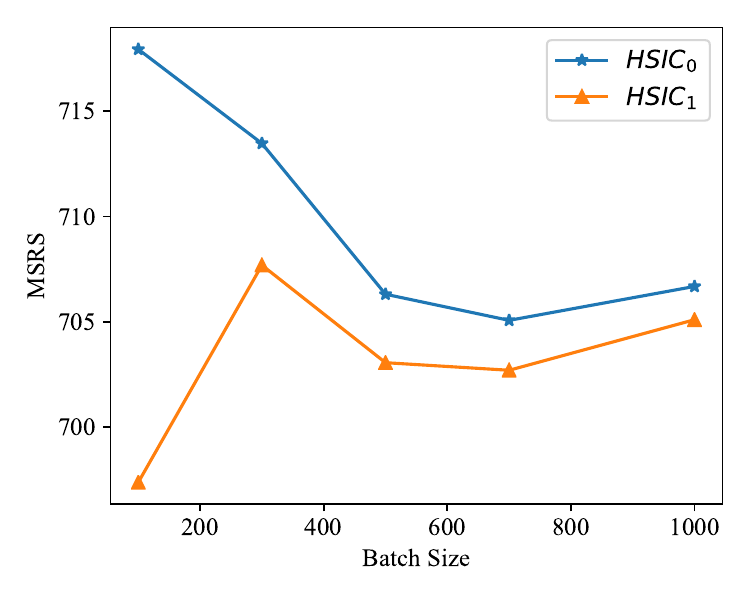}
   \end{minipage}
   \label{fig:HSICa}
 }
 \quad
 \subfloat[]{
   \begin{minipage}[c]{0.21\textwidth}
     \centering
     \includegraphics[width=1\textwidth]{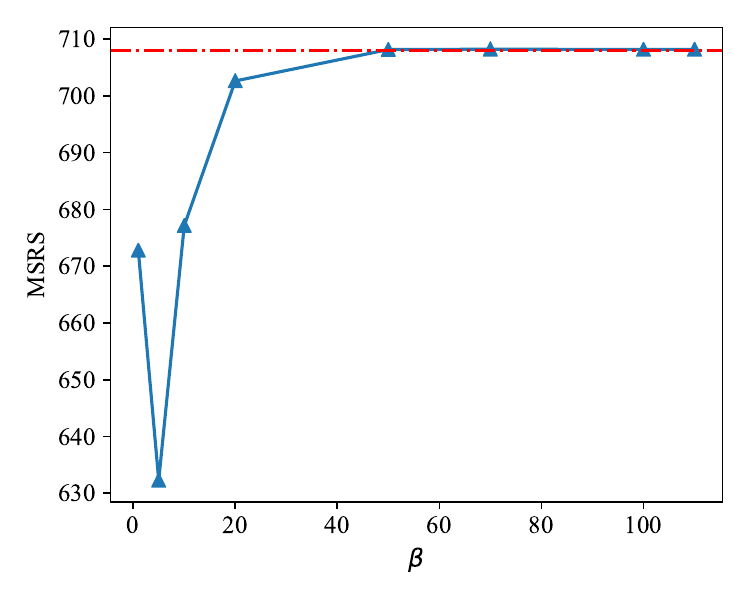}
   \end{minipage}
   \label{fig:HSICb}
 }
 \quad
 \subfloat[]{
   \begin{minipage}[c]{0.21\textwidth}
     \centering
     \includegraphics[width=1\textwidth]{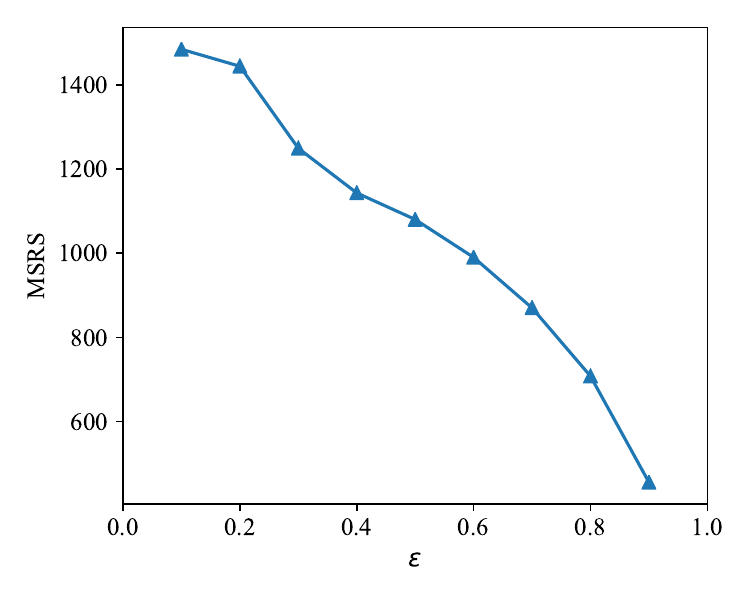}
   \end{minipage}
   \label{fig:HSICc}
 }
 \quad
 \subfloat[]{
   \begin{minipage}[c]{0.21\textwidth}
     \centering
     \includegraphics[width=1\textwidth]{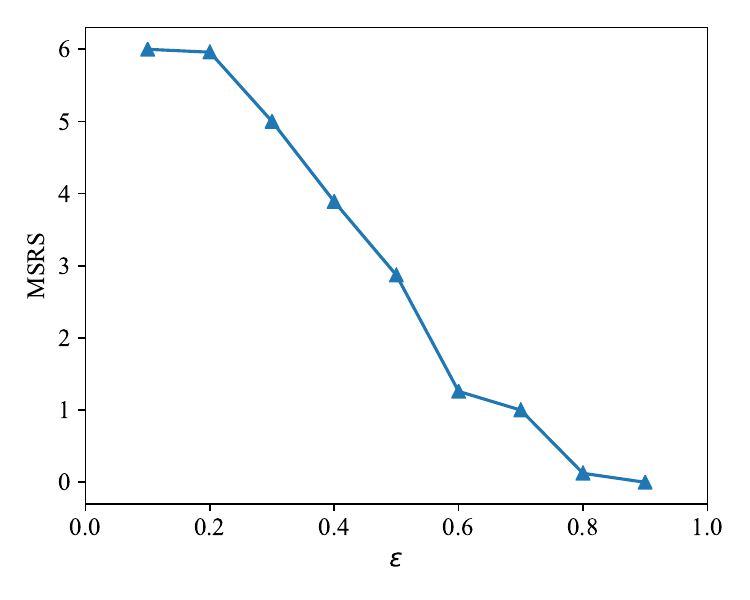}
   \end{minipage}
   \label{fig:HSICd}
 }
\caption{Results for different settings of HSIC, $\beta$ and $\epsilon$.}
\label{fig:HSIC}
\end{figure}

\subsection{Hyperparameter Selection of MSRS}
\label{Hyperparameter selection of MSRS}
 Batch size $n$, scaling factor $\beta$ and model-specific threshold $\epsilon$ are three configurable parameters in calculating MSRS. If not specified, we utilize minibatch of size $n = 256$ and $\beta=100$ to calculate MSRS. As for model-specific threshold $\epsilon$, we set $\epsilon=0.7$ for models of plain structure (e.g., VGGs) and $\epsilon=0.8$ for models of block structure (e.g., ResNets). In this section, we aim to discuss the effect of them on MSRS. 

Since $HSIC_0$ is a biased estimator with $O(\frac{1}{n})$ bias. In order to reduce the impacts of batch size on 
the value of MSRS,
we utilize an unbiased estimator of $HSIC_1$ in place of $HSIC_0$. It can be reasonable to question whether $HSIC_1$ is independent of the batch size. Therefore, we conduct experiments with ResNet110 on CIFAR10 with configurations of $n = 100$, $300$, $500$, $700$, $1,000$ and $\beta=100$. \cref{fig:HSICa} presents the relationship between batch size and MSRS calculated by $HSIC_0$/$HSIC_1$. In contrast to MSRS calculated by $HSIC_0$, MSRS calculated by $HSIC_1$ is less sensitive to changes in $n$ (the standard deviation is $3.4$ for MSRS calculated by $HSIC_1$ while $5.0$ for MSRS calculated by $HSIC_0$). Since batch size has a negligible effect on MSRS calculated by $HSIC_1$, in this paper we uniformly take $n=256$.

As for scaling factor $\beta$, we utilize $\beta = 1$, $5$, $10$, $20$, $50$, $70$, $100$, $110$ (If $\beta > 110$, the result will be out of range for floating point representation in Python\footnote{https://www.python.org}, so we take $\beta$ as maximum as $110$) and $n = 256$ to conduct experiments with ResNet110 on CIFAR10. As shown in \cref{fig:HSICb}, when $\beta \geqslant 50$, MSRS tends to be stable. So, empirically it is indeed possible to fix $\beta$ to a certain value $\geqslant 50$. In this work, we utilize $\beta =50$ for the sake of uniformity.


As for model-specific threshold $\epsilon$, we conducted experiments with ResNet110 on CIFAR10, using $\beta=100$ 
, changing $\epsilon$ from 0.1 to 0.9. As shown in \cref{fig:HSICc}, MSRS is indeed sensitive (inversely proportional) to $\epsilon$. Hence, the selection of $\epsilon$ is critical for calculating MSRS. Next, we introduce a heuristic method to choose $\epsilon$. Due to the space limitation, we only introduce the selection of $\epsilon$ for models of block structure here. First, we choose a well-trained ResNet8 on CIFAR10 with $85.41\%$ test accuracy. Obviously, ResNet8 is underfitting on CIFAR10 and we believe it has no model structural redundancy. Afterwards, we use different $\epsilon$ with $\beta=100$, $n=256$ to calculate MSRS, which is shown in \cref{fig:HSICd}. We obverse that when $\epsilon=0.8$, MSRS is close to 0, and thus utilize it for models of block structure.

\section{Related Work}
\label{sec:related}

\textbf{Representational Similarity Metrics:} Representational similarity computes the similarity between statistics on two different IRs, which gives insight into the interaction between machine learning algorithms and data than the value of the loss function alone. There exist several statistical measures, each of which serves as a different notion of similarity. Hardoon et al.~\cite{hardoon2004canonical} propose canonical correlation analysis (CCA), which aims to find basis vectors such that the correlation between the projections of variables onto these basis vectors is mutually maximized. Since CCA is sensitive to perturbation, Raghu et al.~\cite{raghu2017svcca} and Morcos et al. \cite{morcos2018insights} propose Singular Vector CCA and Projection Weighted CCA to reduce the sensitivity of CCA to perturbation, respectively. However, these metrics are invariant to invertible linear transformation and can not measure meaningful similarities between IRs of higher dimensions than the number of data points. To address the above challenges, Kornblith et al. \cite{kornblith2019similarity} introduce CKA to measure similarities between IRs. Nguyen et al.~\cite{DBLP:conf/iclr/NguyenRK21} investigate the effects of depth and width on IRs on the basis of CKA, and find that overly complicated models exhibit the block structure. Different from these works, which purely compute the representational similarity, we associate representational similarity with model structural redundancy. Furthermore, we explore the redundancy issues in state-of-the-art DNNs.

\textbf{Layer Pruning:} The core of layer pruning lies in the selection of layers, which should achieve the lowest compromise in accuracy with the highest compression ratio. Chen et al. \cite{chen2018shallowing} fit a linear classifier to evaluate the performance of a layer thus finding the ones to be removed. On the basis of \cite{chen2018shallowing}, Elkerdawy et al. \cite{elkerdawy2020one} utilize imprinting to efficiently approximate the accuracy of each layer and prune the layer which achieves minimal accuracy improvement compared to the preceding layer. Zhou et al. \cite{zhou2021evolutionary} reconstruct the loss of layer pruning as a dual-objective function that minimizes the error and number of blocks, and further adopts multi-objective evolutionary algorithm to solve it. Wang et al. \cite{wang2021accelerate} formalize the relationships between accuracy and depth/width/resolution as a polynomial regression and obtain the optimal values for these dimensions by solving the polynomial regression. De et al. \cite{de2022depth} replace the last few layers with an auxiliary network as an effective interpreter of IRs to produce the pruned model. Unlike the existing works, we propose a redundancy-aware layer pruning method with the guidance of layer similarity to remove those redundant layers, which has been demonstrated to be superior to existing works by extensive experimental results.

\textbf{NAS:} 
Since the state of the arts, e.g., VGG \cite{DBLP:journals/corr/SimonyanZ14a} and ResNet \cite{he2016deep} are manually designed by experts by a trial-and-error process, which requires substantial resources and time even for experts. To cope with the aforementioned challenges, NAS, as a technique for automating the architecture designs of DNNs is proposed to reduce onerous development costs. For example, \cite{DBLP:conf/aaai/RealAHL19,pmlr-v70-real17a,DBLP:journals/corr/abs-2008-10937,DBLP:conf/iccv/XieY17} leverage evolutionary algorithms to search for novel architectures automatically. Besides, Zoph et al. \cite{DBLP:conf/iclr/ZophL17} utilize a recurrent neural network (RNN) as a controller to compose neural network architectures and train this RNN with reinforcement learning to obtain optimal architectures on a validation set. Then many follow-up methods \cite{DBLP:conf/iclr/BakerGNR17,Zoph_2018_CVPR,Zhong_2018_CVPR,pmlr-v80-pham18a} improve this framework with different controller policies and neural architecture encoding. Compared to the above-mentioned studies sampling architectures from a discrete search space, some works \cite{DBLP:conf/iclr/LiuSY19,DBLP:journals/corr/abs-1909-06035,DBLP:conf/iccv/ChenXW019} search over a continuous and differentiable search space, which can be optimized with gradient descent. Generally speaking, NAS aims to search for the optimal architecture for a given resource budget that achieves an accuracy-budget trade-off. Unfortunately, as we mentioned in our paper, existing resource budgets (e.g., Latency) can not quantitatively measure the remaining redundancy in DNN models. To bridge this gap, we propose MSRS to quantitatively assess the degree of redundancy in the model structure, and further present a redundancy-aware NAS algorithm to search for the optimal architecture.

\section{Discussion and Threats to Validity}


This subsection summarizes the findings and insights obtained in this paper and followed by the threats to validity of our work.

\textbf{Discussion.} This paper proposes a testing metric called MSRS to quantitatively measure the degree of redundancy in DNN models. With MSRS, we systematically investigate the redundancy issue in DNN models and confirm the ubiquitous presence of model redundancy, sometimes even to a surprisingly high level, which urgently calls for model structure optimization. Equipped with MSRS, we further present a novel testing framework \rt, with the aim to facilitate further studies on optimizing model structures in a more computationally effective environment. Our large-scale study marks the first step towards such a direction, with interesting findings identified.

\textbf{Threats to validity.} The selection of subject datasets and models could be a threat to validity. We try to counter this by using four publically available datasets with diverse scales, together with a large collection of widely used DNN models that achieve competitive performance.
A further threat would be the randomness factors for computing MSRS. We try to counteract the effects potentially brought by randomness by repeating each configuration 10 times and reporting the averaged results.

\begin{figure}[hbpt]
  \centering
   \includegraphics[width=1\linewidth]{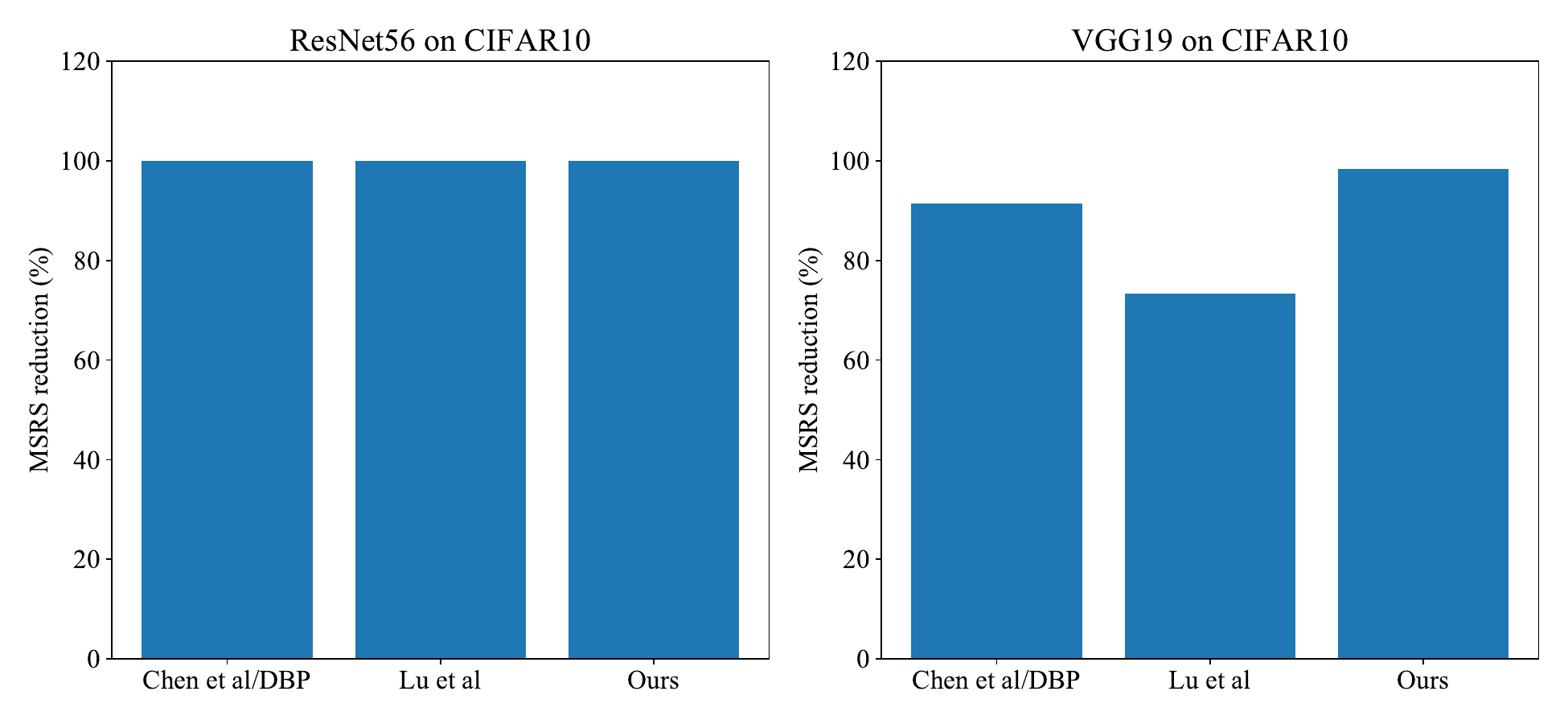}
   \caption{MSRS reduction after layer pruning.}
   \label{fig:msrs_before_after_pruning}
\end{figure}

\begin{figure}[htbp]
  \centering
   \includegraphics[width=0.7\linewidth]{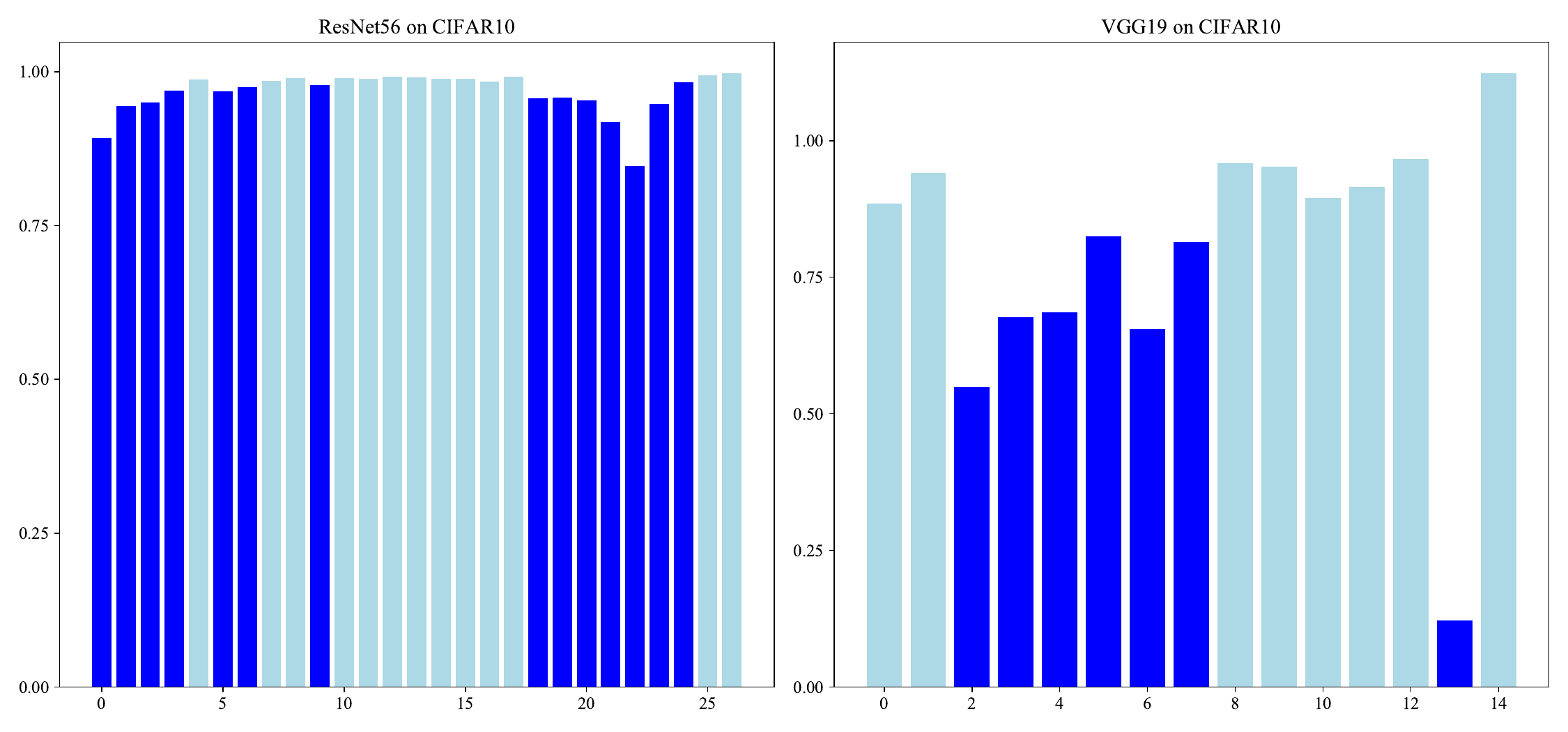}
   \caption{CKA values of two adjacent layers. The bars with the transparent color represent that the current layers are similar to their front layer and are considered redundant.}
   \label{fig:CKA pruning}
  \vspace{-4mm}
\end{figure}

\begin{table}[htbp]
\centering
\scalebox{0.9}{\begin{tabular}{|c|c|c|c|c|}
\hline
Datasets       & Image Size  & Training & Testing & Classes \\ \hline
CIFAR10~\cite{krizhevsky2009learning}        & 32, 32, 3   & 50K      & 10K     & 10      \\ \hline
CIFAR100~\cite{krizhevsky2009learning}       & 32, 32, 3   & 50K      & 10K     & 100     \\ \hline
ImageNet~\cite{DBLP:journals/ijcv/RussakovskyDSKS15}       & 224, 224, 3 & 1.28M    & 50K     & 1000    \\ \hline
ImageNet16-120~~\cite{chrabaszcz2017downsampled} & 16, 16, 3   & 151.7K   & 3K      & 120     \\ \hline
\end{tabular}}
\label{tab:datasets}
\caption{Datasets characteristics.}
\vspace{-3mm}
\end{table}

\begin{table}[htbp]
\centering
\scalebox{0.9}{\begin{tabular}{|c|c|c|c|c|c|}
\hline
Dataset        & Number of models & $T$ & $M$ & $w$ & $\lambda$ \\ \hline
CIFAR10        & 11715            & 0.01             & 35            & -0.20         & 0.1                    \\ \hline
CIFAR100       & 11748            & 0.01             & 35            & -0.24         & 0.2                    \\ \hline
ImageNet16-120 & 11741            & 0.01             & 35            & -0.20         & 0.5                    \\ \hline
\end{tabular}}
\caption{Hyperparameter settings of redundancy-aware NAS using Latency.}
\label{tab:addtional hyper}%
\vspace{-3mm}
\end{table}

\begin{table}[htbp]
\centering
\scalebox{0.9}{\begin{tabular}{|c|c|c|c|c|c|}
\hline
Dataset        & Number of models & $P$ & $M$ & $w$ & $\lambda$ \\ \hline
CIFAR10        & 11715            & 0.15             & 35            & -0.035         & 0.4                    \\ \hline
CIFAR100       & 11748            & 0.15              & 35            & -0.06         & 0.8                    \\ \hline
ImageNet16-120 & 11741            & 0.15              & 35            & -0.073         & 0.2                    \\ \hline
\end{tabular}}
\caption{Hyperparameter settings of redundancy-aware NAS using Params.}
\label{tab:addtional hyper for Params}%
\vspace{-3mm}
\end{table}

\begin{table}[htbp]
\centering
\scalebox{0.9}{\begin{tabular}{|c|c|c|c|c|c|}
\hline
Dataset        & Number of models & $F$ & $M$ & $w$ & $\lambda$ \\ \hline
CIFAR10        & 11715            & 10               & 35            & -0.03         & 0.15                    \\ \hline
CIFAR100       & 11748            & 10               & 35            & -0.06         & 0.4                    \\ \hline
ImageNet16-120 & 11741            & 10               & 35            & -0.073         & 0.6                    \\ \hline
\end{tabular}}
\caption{Hyperparameter settings of redundancy-aware NAS using FLOPs.}
\label{tab:addtional hyper for FLOPs}%
\end{table}

\begin{table}[htbp]
\centering
\scalebox{0.5}{\begin{tabular}{|c|c|c|c|c|c|}
\hline
Model                     & Method         & Top-1\% & Params(PR) & FLOPs(PR) & Latency(PR) \\ \hline
\multirow{4}{*}{ResNet56} & Original       & 71.81   & 0\%        & 0\%       & 0\%         \\ \cline{2-6} 
                          & Chen et al.~\cite{chen2018shallowing}/DBP~\cite{wang2019dbp} & $70.45 \pm 0.18$   & 36.67\%    & 41.39\%   & 39.53\%     \\ \cline{2-6} 
                          & Lu et al.~\cite{lu2021graph}       & $69.24 \pm 0.10$   & 10.83\%    & 41.65\%   & 37.94\%     \\ \cline{2-6} 
                          & Ours           & $70.09 \pm 0.10$   & 31.32\%    & 48.99\%   & 45.78\%     \\ \hline
\multirow{4}{*}{VGG19}    & Original       & $72.63$   & 0\%        & 0\%       & 0\%         \\ \cline{2-6} 
                          & Chen et al.~\cite{chen2018shallowing}/DBP~\cite{wang2019dbp} & $69.73 \pm 0.15$   & 71.81\%    & 45.00\%   & 40.70\%     \\ \cline{2-6} 
                          & Lu et al.~\cite{lu2021graph}       & $66.88 \pm 0.25$   & 41.19\%    & 45.00\%   & 40.35\%     \\ \cline{2-6} 
                          & Ours           & $70.33 \pm 0.44$   & 54.63\%    & 52.00\%   & 38.60\%     \\ \hline
\end{tabular}}
\caption{Experimental results on the CIFAR100 dataset using VGG19 and ResNet56, PR is the pruning rate.}
\label{tab:CKA pruning on cifar100}
\end{table}


\begin{figure*}[htbp]
\centering
 \subfloat[]{
   \begin{minipage}[c]{0.31\textwidth}
     \centering
     \includegraphics[width=0.99\textwidth]{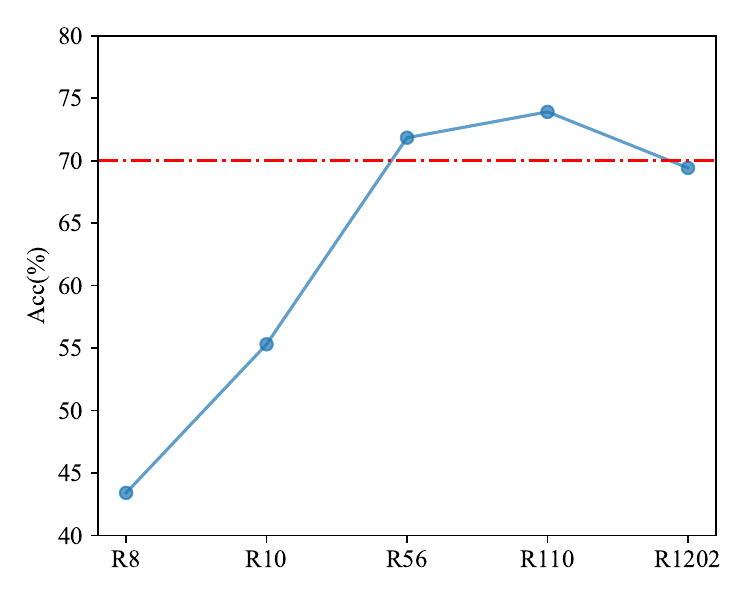}
   \end{minipage}
   \label{fig:over_under_fitting1_cifar100}
 }
 \subfloat[]{
   \begin{minipage}[c]{0.31\textwidth}
     \centering
     \includegraphics[width=0.99\textwidth]{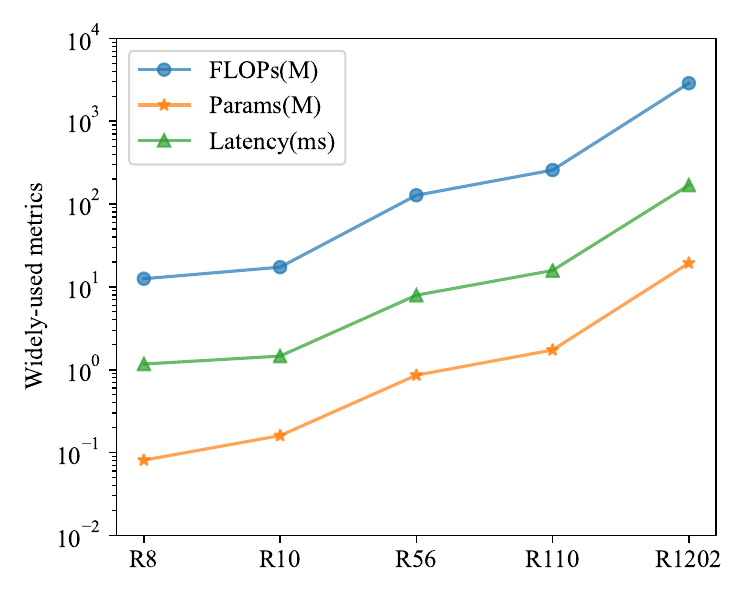}
   \end{minipage}
   \label{fig:over_under_fitting2_cifar100}
 }
 \subfloat[]{
   \begin{minipage}[c]{0.31\textwidth}
     \centering
     \includegraphics[width=0.99\textwidth]{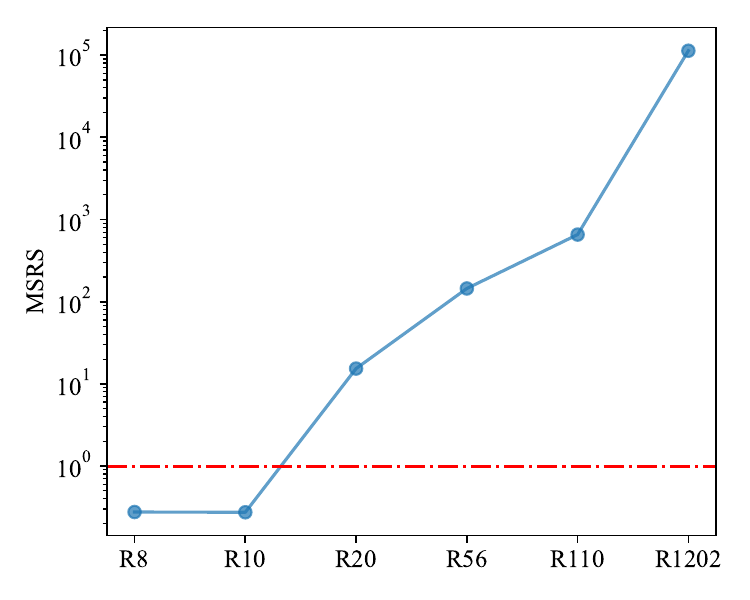}
   \end{minipage}
   \label{fig:over_under_fitting3_cifar100}
 }
\caption{Differences between existing metrics and MSRS for a family of ResNets on CIFAR100. R denotes ResNet.}
\label{fig:over_under_fitting_cifar100}
\end{figure*}

\begin{table*}[htbp]
\begin{threeparttable}
\scalebox{0.9}{\begin{tabular}{|c|cccc|cllll|}
\hline
\multicolumn{1}{|l|}{\multirow{2}{*}{}} & \multicolumn{4}{c|}{VGGs}                                                                            & \multicolumn{5}{c|}{ResNets\tnote{1}}                                                                                                                                                         \\ \cline{2-10} 
\multicolumn{1}{|l|}{}                  & \multicolumn{1}{c|}{VGG11}   & \multicolumn{1}{c|}{VGG13}   & \multicolumn{1}{c|}{VGG16}   & VGG19   & \multicolumn{1}{c|}{ResNet20(18)} & \multicolumn{1}{c|}{ResNet32(34)} & \multicolumn{1}{c|}{ResNet44(50)} & \multicolumn{1}{c|}{ResNet56(101)} & \multicolumn{1}{c|}{ResNet110(152)} \\ \hline
CIFAR10                                 & \multicolumn{1}{c|}{92.11\%} & \multicolumn{1}{c|}{93.68\%} & \multicolumn{1}{c|}{93.63\%} & 93.36\% & \multicolumn{1}{c|}{91.78\%}             & \multicolumn{1}{c|}{92.38\%}             & \multicolumn{1}{c|}{92.92\%}             & \multicolumn{1}{c|}{93.11\%}              & \multicolumn{1}{c|}{93.78\%}                                    \\ \hline
CIFAR100                                & \multicolumn{1}{c|}{66.87\%} & \multicolumn{1}{c|}{70.19\%} & \multicolumn{1}{c|}{72.34\%} & 72.63\% & \multicolumn{1}{c|}{58.61\%}             & \multicolumn{1}{c|}{69.78\%}             & \multicolumn{1}{c|}{71.32\%}             & \multicolumn{1}{c|}{71.81\%}              & \multicolumn{1}{c|}{73.91\%}                                    \\ \hline
ImageNet                              & \multicolumn{1}{c|}{70.36\%} & \multicolumn{1}{c|}{71.55\%} & \multicolumn{1}{c|}{73.48\%} & 74.17\% & \multicolumn{1}{c|}{69.67\%}      & \multicolumn{1}{c|}{73.22\%}      & \multicolumn{1}{c|}{75.99\%}      & \multicolumn{1}{c|}{77.32\%}       & \multicolumn{1}{c|}{78.26\%}        \\ \hline
\end{tabular}}
\begin{tablenotes}   
\footnotesize           
\item[1] We train ResNet20, 32, 44, 56, 110 on CIFAR10 and CIFAR100, ResNet18, 34, 50, 101, 152 on ImageNet.        
\end{tablenotes}      
\end{threeparttable}  
\caption{Models Performance.}
\label{Models statistics}
\end{table*}


\end{document}